\documentclass{article}

\usepackage{PRIMEarxiv}

\usepackage[utf8]{inputenc} 
\usepackage[T1]{fontenc}    
\usepackage{hyperref}       
\usepackage{url}            
\usepackage{booktabs}       
\usepackage{amsfonts}       
\usepackage{nicefrac}       
\usepackage{microtype}      
\usepackage{lipsum}
\usepackage{fancyhdr}       
\usepackage{graphicx}       
\graphicspath{{media/}}     
\usepackage{subcaption}
\title{STAResNet: a Network in Spacetime Algebra to solve Maxwell's PDEs
\thanks{\textit{\underline{Citation}}: 
\textbf{Authors. Title. Pages.... DOI:000000/11111.}} 
}

\author{
  Alberto Pepe, Joan Lasenby \\
  Signal Processing and Communications Lab \\
  University of Cambridge \\
  Cambridge, UK\\
  \texttt{\{ap2219, jl221\}@cam.ac.uk} \\
   \And
  Sven Buchholz \\
  Department of Computer Science and Media \\
  Technische Hochschule Brandenburg\\
  University of Applied Sciences \\
  Brandenburg, Germany\\
  \texttt{sven.buchholz@th-brandenburg.de} \\
}

\begin{document}
\maketitle

\begin{abstract}
We introduce STAResNet, a ResNet architecture in Spacetime Algebra (STA) to solve Maxwell's partial differential equations (PDEs). Recently, networks in Geometric Algebra (GA) have been demonstrated to be an asset for truly geometric machine learning. In \cite{brandstetter2022clifford}, GA networks have been employed for the first time to solve partial differential equations (PDEs), demonstrating an increased accuracy over real-valued networks. In this work we solve Maxwell's PDEs both in GA and STA employing the same ResNet architecture and dataset, to discuss the impact that the choice of the right algebra has on the accuracy of GA networks. Our study on STAResNet shows how the correct geometric embedding in Clifford Networks gives a mean square error (MSE), between ground truth and estimated fields, up to 2.6 times lower than than obtained with a standard Clifford ResNet with 6 times fewer trainable parameters. STAREsNet demonstrates consistently lower MSE and higher correlation regardless of scenario. The scenarios tested are: sampling period of the dataset; presence of obstacles with either seen or unseen configurations;  the number of channels in the ResNet architecture; the number of rollout steps; whether the field is in 2D or 3D space. This demonstrates how choosing the right algebra in Clifford networks is a crucial factor for more compact, accurate, descriptive and better generalising pipelines. 
\end{abstract}


\section{Introduction}

Geometric Algebra Networks, also known as Clifford Algebra Networks, leverage the mathematical framework of Geometric Algebra to represent and manipulate data. Geometric Algebra is a powerful, high-dimensional algebraic system that extends traditional linear algebra, enabling the compact and intuitive representation of geometric transformations, rotations, and reflections \cite{doran2003geometric,lasenby2011guide,hitzer2012introduction}. In GA networks, data and operations are expressed in terms of multivectors, which can capture complex geometric relationships more naturally than traditional tensor or matrix representations.

Early proposals for neural networks working in Geometric Algebra (GA) can be found in the literature from the end of the last century \cite{pearson1994neural, bayro1997geometric, pearson2003clifford}. However, it is only in the past few years that the need for an effective and intuitive approach to geometrical problems in learning has sparked renewed interest in the field. Today, several architectures in GA exist, capable of handling convolutions and Fourier transforms \cite{brandstetter2022clifford, pepe2024garelu}, performing rotations and rigid body motions \cite{pepe2024cgaposenet+, ruhe2023geometric}, and preserving end-to-end equivariance \cite{ruhe2024clifford, pepe2024clifford, brehmer2024geometric}.

In recent years, the use of machine learning, particularly neural networks, to solve partial differential equations (PDEs) has gained significant traction \cite{long2018pde,liang2024solving, karniadakis2021physics,zhang2021mod}. State-of-the-art approaches include physics-informed neural networks (PINNs) \cite{cai2021physics1,mao2020physics,cai2021physics2}, Fourier neural operators \cite{li2020fourier,bonev2023spherical}, and deep Ritz methods \cite{uriarte2023deep, yu2018deep}. These methods aim to approximate solutions to PDEs by embedding the physical laws described by the PDEs directly into the training process of neural networks. For example, PINNs incorporate the residuals of the PDEs into the loss function, ensuring that the neural network's output satisfies the underlying physical laws.

When solving PDEs, GA networks offer several unique advantages. One of the primary benefits is their ability to preserve the geometric meaning of data. By embedding data in the correct algebraic framework, GA networks can handle geometric transformations and physical laws more naturally. This is particularly advantageous for problems involving rotations, reflections, and other geometric transformations, which are common in physics and engineering. Moreover, GA networks can represent complex multi-dimensional relationships compactly and efficiently, making them well-suited for high-dimensional PDEs. The ability to handle convolutions and Fourier transforms within the GA framework \cite{brandstetter2022clifford} also enables GA networks to process and analyze spatial and frequency domain information effectively, which is crucial for many PDE problems.

The question we address here is: how does the choice of the algebra in which data is embedded affect the accuracy of the PDE solution via a GA network? We focus on Maxwell's PDEs, which describe the fundamental behavior of electric and magnetic fields, and compare their solutions using two approaches: one in $n$D GA and one in $(n+1)$D Spacetime Algebra (STA).

By comparing these two approaches, we aim to understand the impact of the algebraic framework on the accuracy and efficiency of the solutions. The choice of algebra can significantly influence the complexity of the problem formulation and the performance of the neural network, potentially leading to more accurate and efficient PDE solvers.

\section{Problem Definition}

axwell's equations describe the behaviour of electric and magnetic fields in classical electromagnetism. They read as follows (for the vacuum case):
\begin{equation}
\nabla \cdot \mathbf{E} = \frac{\rho}{\varepsilon_0}
\label{m1}
\end{equation}
known as \textbf{Gauss's Law for Electricity} and states that the divergence of the electric field (\(\mathbf{E}\)) is equal to the charge density (\(\rho\)) divided by the permittivity of free space (\(\varepsilon_0\)).

\begin{equation}
\nabla \cdot \mathbf{B} = 0
\label{m2}
\end{equation}
representing \textbf{Gauss's Law for Magnetism} and indicating that the divergence of the magnetic field (\(\mathbf{B}\)) is zero, implying the absence of magnetic monopoles.

\begin{equation}
\nabla \times \mathbf{E} +\frac{\partial \mathbf{B}}{\partial t} = 0
\label{m3}
\end{equation}
i.e. \textbf{Faraday's Law of Induction}, stating that the curl of the electric field (\(\mathbf{E}\)) is equal to the negative rate of change of the magnetic field (\(\mathbf{B}\)) with respect to time. 

\begin{equation}
\nabla \times \mathbf{B}  - \mu_0 \varepsilon_0 \frac{\partial \mathbf{E}}{\partial t} = \mu_0 \mathbf{J}
\label{m4}
\end{equation}
which is the \textbf{Ampère's Law}, stating that the curl of the magnetic field (\(\mathbf{B}\)) is equal to the sum of the current density (\(\mathbf{J}\)) and the product of the permeability of free space (\(\mu_0\)) and the rate of change of the electric field (\(\mathbf{E}\)) with respect to time.  To simplify the exposition we will work in natural units ($c=\epsilon_0 = \mu_0 = 1$), so that Maxwell's equations become:

\begin{equation}
    \nabla \cdot \mathbf{E}  =  {\rho}  \;\;\;\;\;\;  \nabla \cdot \mathbf{B} = 0  \;\;\;\;\;\;
    \nabla \times \mathbf{E}    =  -\frac{\partial \mathbf{B}}{\partial t}  \;\;\;\;\;\;   \nabla \times \mathbf{B}  = \frac{\partial \mathbf{E}}{\partial t} + \mathbf{J}
\end{equation}

\subsection{Maxwell's equations in G(3,0,0)}

This scenario will be our baseline, and it is equivalent to that proposed in \cite{brandstetter2022clifford}, using Euclidean 3-space. G(3,0,0) is the three-dimensional Geometric Algebra spanned by the three basis vectors $\{e_1, e_2, e_3\}$. By defining the pseudoscalar $i = e_{123}$ we can introduce the electromagnetic (EM) field multivector $F$:

\begin{equation}
    F = \mathbf{E} + i\mathbf{B} =  E_1 e_1 +E_2 e_2 +E_3 e_3 + B_1 e_{23} + B_2 e_{13} + B_3 e_{12}.  
\end{equation} 

With this object can pair Eq. \ref{m1}-\ref{m4} into a set of two equations only: \begin{equation}
    \nabla \cdot F = {\rho} \;\;\;\;\;\;
    \nabla \times F = i\left(\frac{\partial F}{\partial t} + \mathbf{J}\right)
\end{equation}

which, using the geometric product, can be further reduced to:

\begin{equation}
   \left( \frac{\partial}{\partial t} + i\nabla \right) F =   \mathbf{J} - i\rho
\end{equation}

\begin{figure}[!htbp]
    \centering
    \includegraphics[width=0.65\textwidth]{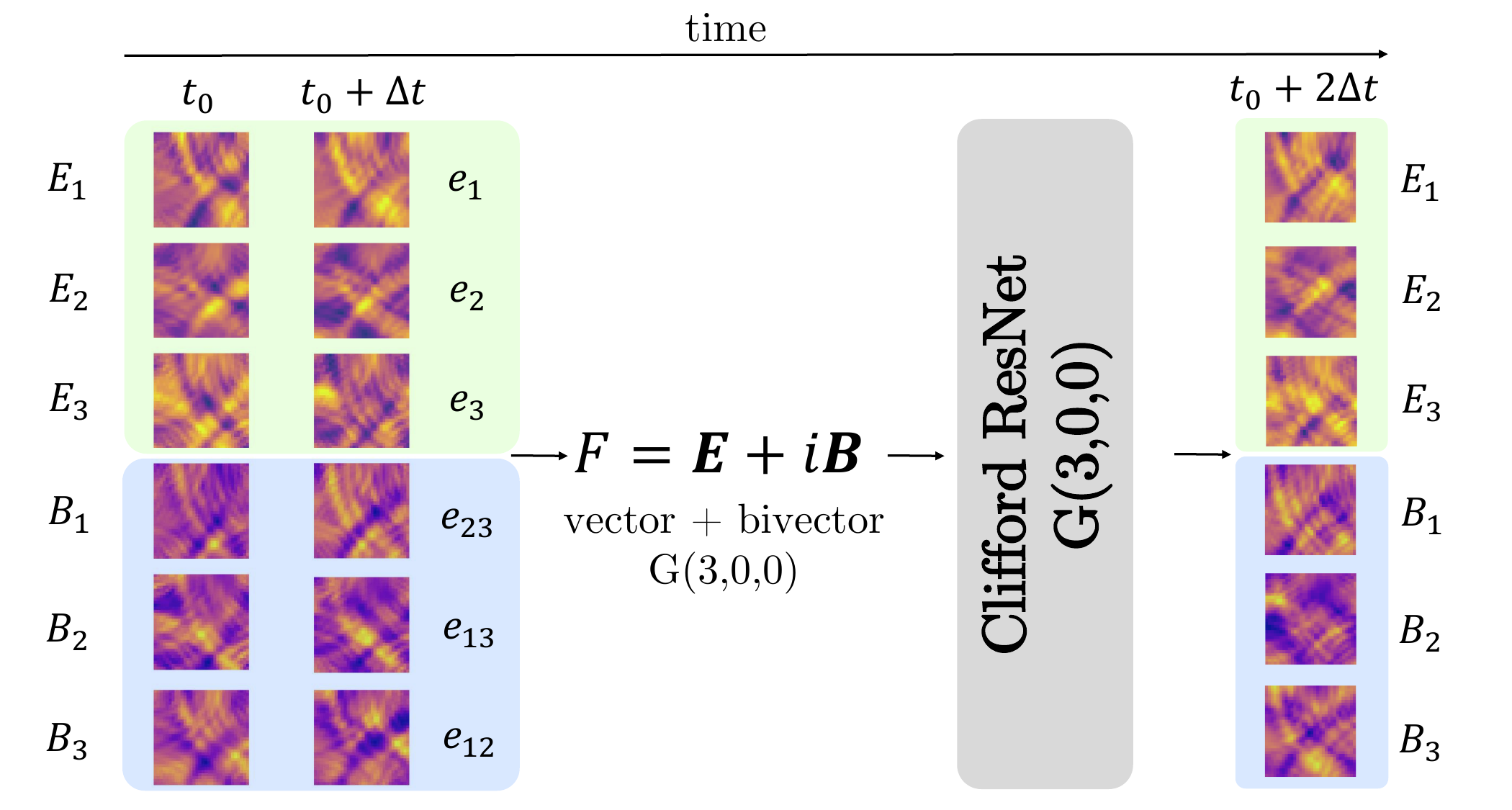}
    \caption{3D GA approach: solving Maxwell's PDEs through Clifford ResNet, a ResNet-inspired network working in G(3,0,0).}
    \label{fig:resnetg300}
\end{figure}

\vskip 0.1cm
In this scenario, we formulate the PDE solution as a 3D multivector-to-multivector regression problem. The inputs to the network will be a pair of multivectors sampled at two consecutive time instants $\{ F_{i}, F_{i + \Delta t}\}$, while the label will be the multivector after a history of 2 time steps, i.e. $F_{i + 2\Delta t}$. This is summarized in Fig. \ref{fig:resnetg300}.  

\subsection{Maxwell's equations in G(1,3,0)}

However, the previous section chose an embedding with little physical meaning. We know that if we choose the Spacetime Algebra (STA) Maxwell's equations have a much more natural embedding. The STA is G(1,3,0), in which we have four basis vectors $\{ \gamma_{\mu}\}$, with $\gamma_0 ^ 2 = - \gamma_k ^2 = 1$ and $k = 1, 2, 3$.  In STA, the electric and magnetic fields are both bivectors of the form: 

\begin{equation}
    \mathbf{E} = E_1 \sigma_1 + E_2 \sigma_2 + E_3 \sigma_3
\end{equation}
\begin{equation}
    \mathbf{B} = B_1 \sigma_1 + B_2 \sigma_2 + B_3 \sigma_3
\end{equation}

\vskip 0.1cm
with $\sigma_k = \gamma_k \gamma_0$. In STA our gradient operator $\nabla$ includes the time derivative, ie 

\[ \nabla = \gamma^{i} \frac{\partial}{\partial x_i} \]

\vskip 0.1cm
where we sum over $i=0,1,2,3$, and the $x_i$ are the coordinates. Using the STA $\nabla$ we are able to write Maxwell's equations in a single equation:

 \begin{equation}
    \nabla F = J
\label{final}
\end{equation} 

where the Faraday bivector $F$ is given by, $F = \mathbf{E} + I\mathbf{B}$ and the spacetime current $J$ is given by $J = (\rho - \mathbf{J})\gamma_0$.

 Eq. \ref{final} is not only an extremely compact rendition of Maxwell's equations, especially if compared to Eq.\ref{m1}-\ref{m4}, but it has also the advantage of being expressed in terms of a geometric product, which is invertible. This means that working in STA offers an easier way of computing the propagation of the EM field in a conducting medium. We wish to show that a neural network for predicting EM fields working in the STA performs better than working in an unnatural embedding  in G(3,0,0): to do this we will keep data and architectures identical.

\begin{figure}[!htbp]
    \centering
    \includegraphics[width=0.65\textwidth]{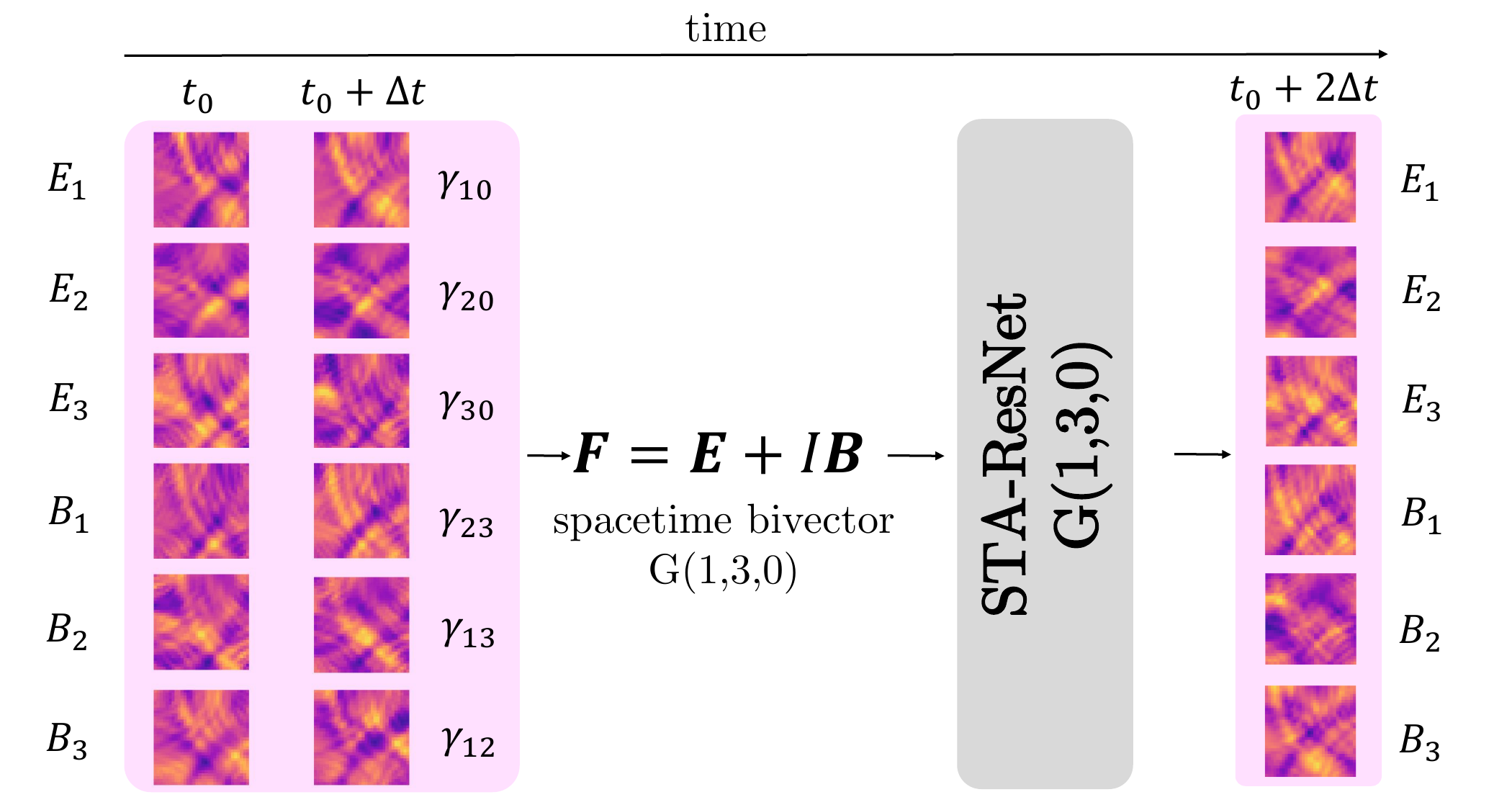}
    \caption{STA approach: solving Maxwell's PDEs through STAResNet, our ResNet-inspired network working in G(1,3,0).}
    \label{fig:resnetg130}
\end{figure}

In this scenario, we formulate the PDEs solution as a spacetime bivector-to-bivector regression problem. The spacetime bivector has form: \begin{equation}
    \mathbf{F} = \mathbf{E} + I\mathbf{B} =  E_1 \gamma_{10} +E_2 \gamma_{20} +E_3 \gamma_{30} + B_1 \gamma_{13} + B_2 \gamma_{13} + B_3 \gamma_{12}.  
    \label{faradaybiv}
\end{equation} The inputs to the network will be a pair of Faraday bivectors sampled at two consecutive time instants $\{ \mathbf{F}_{i}, \mathbf{F}_{i + \Delta t}\}$, while the label will be the bivector after a history of 2 time steps, i.e. $\mathbf{F}_{i + 2\Delta t}$.  The STA approach via STAResNet is summarized in Fig. \ref{fig:resnetg130}. Note how the difference with respect to the approach in 2.2 is exclusively the mathematical framework the network works in.

\section{Approach}

\subsection{Architecture}

We study 2D and 3D Maxwell's PDEs. In the 2D case, we compared the 2D Clifford ResNet and the 2D STAResNet: 2D Clifford ResNet, that handles multivectors $F$ in G(2,0,0), has 20 blocks of 2D convolutions paired with a ReLU activation function. Each convolutional layer has $C = 32$ channels, excepts the first one with $2$ channels and the last one with $1$ channel, corresponding to the number of input and output time steps, respectively. 

2D STAResNet, that handles bivectors $\mathbf{F}$ in G(1,2,0), shares the same structure with the 2D Clifford ResNet, except for the embedding in a different mathematical space and the number of channels, reduced to $C = 24$. This has been done to make sure both networks have the same number of parameters for a fair comparison. Both networks have slightly below 1M parameters.  

In the 3D case, the 3D Clifford ResNet sits in G(3,0,0) while the 3D STAResNet sits in G(1,3,0). The architectures are identical to their 2D counterparts, with the only exceptions that 2D convolutions are replaced with 3D convolutions and the number of channels are reduced from 32 to 11 in Clifford ResNet and from 24 to 8 in STAResNet, for a total of approximately 600,000 parameters for both networks.

\begin{figure*}[!htpb]
        \centering
        \begin{subfigure}[b]{0.45\textwidth}
            \centering
            \includegraphics[width=\textwidth]{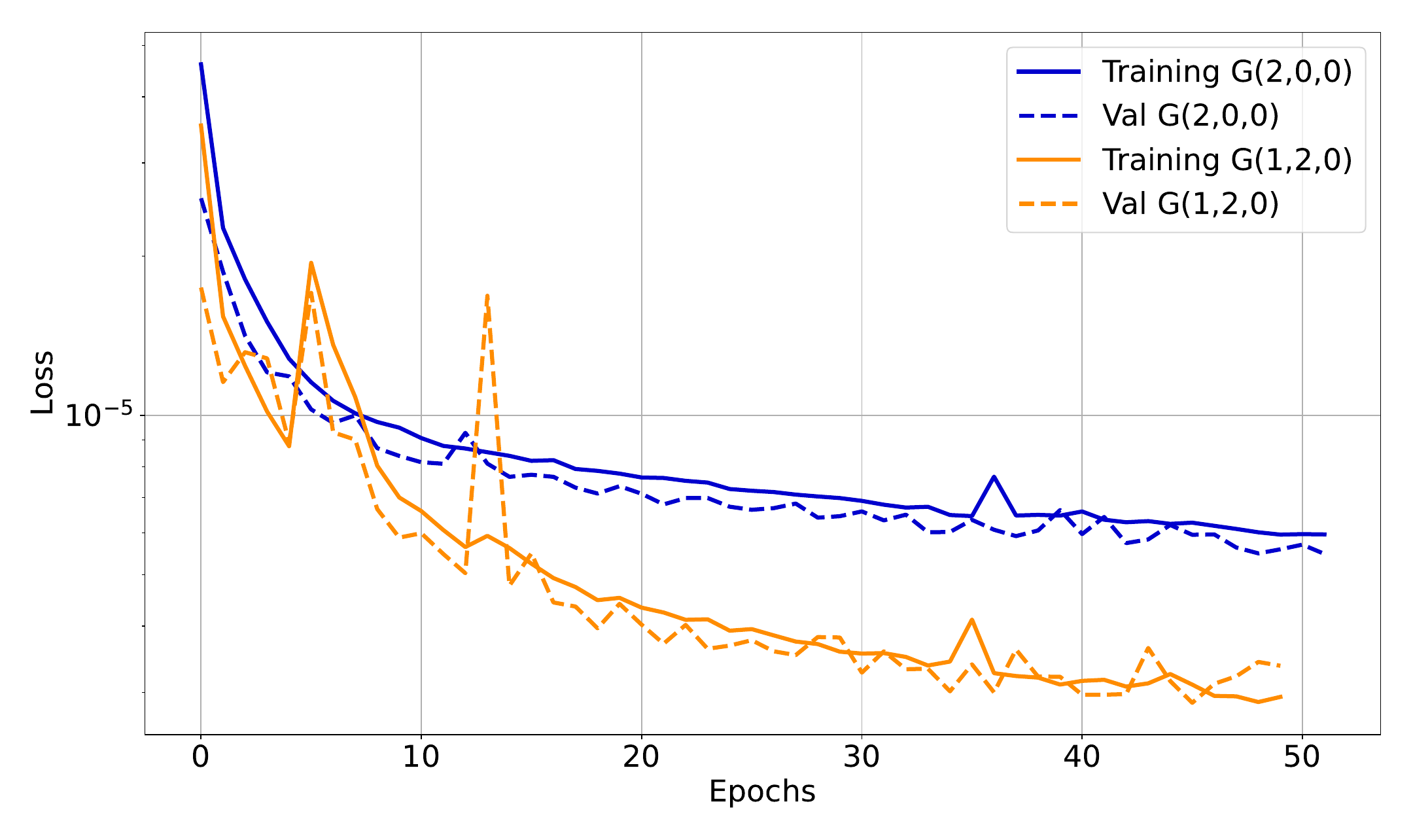}
            \caption[Network2]%
            {{\small $\Delta t = 25$s}}    
            \label{fig:lossesdt25}
        \end{subfigure}
        \hfill
        \begin{subfigure}[b]{0.45\textwidth}  
            \centering 
            \includegraphics[width=\textwidth]{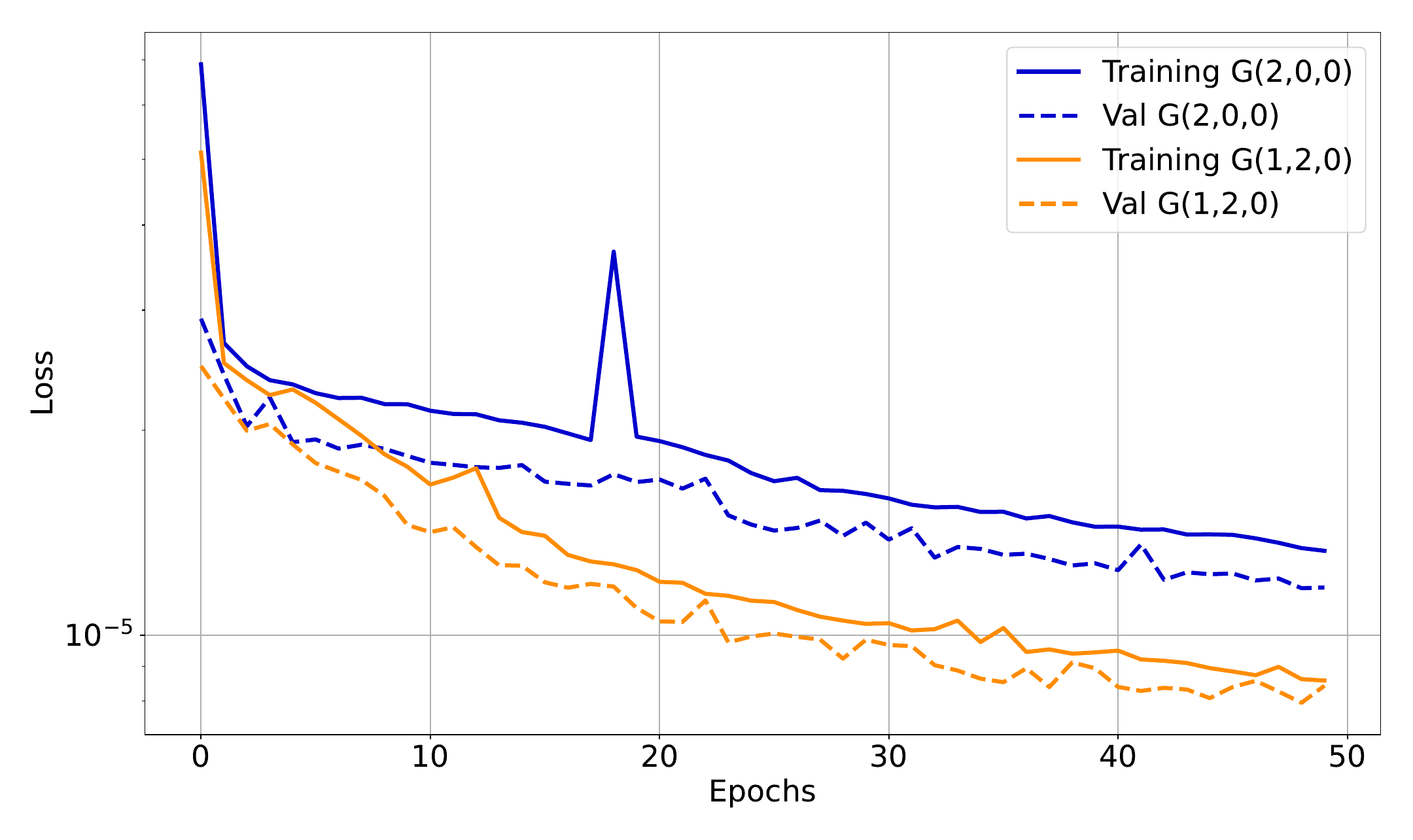}
            \caption[]%
            {{\small $\Delta t = 50$s}}    
            \label{fig:lossesdt50}
        \end{subfigure}
        \vskip\baselineskip
        \begin{subfigure}[b]{0.45\textwidth}   
            \centering 
            \includegraphics[width=\textwidth]{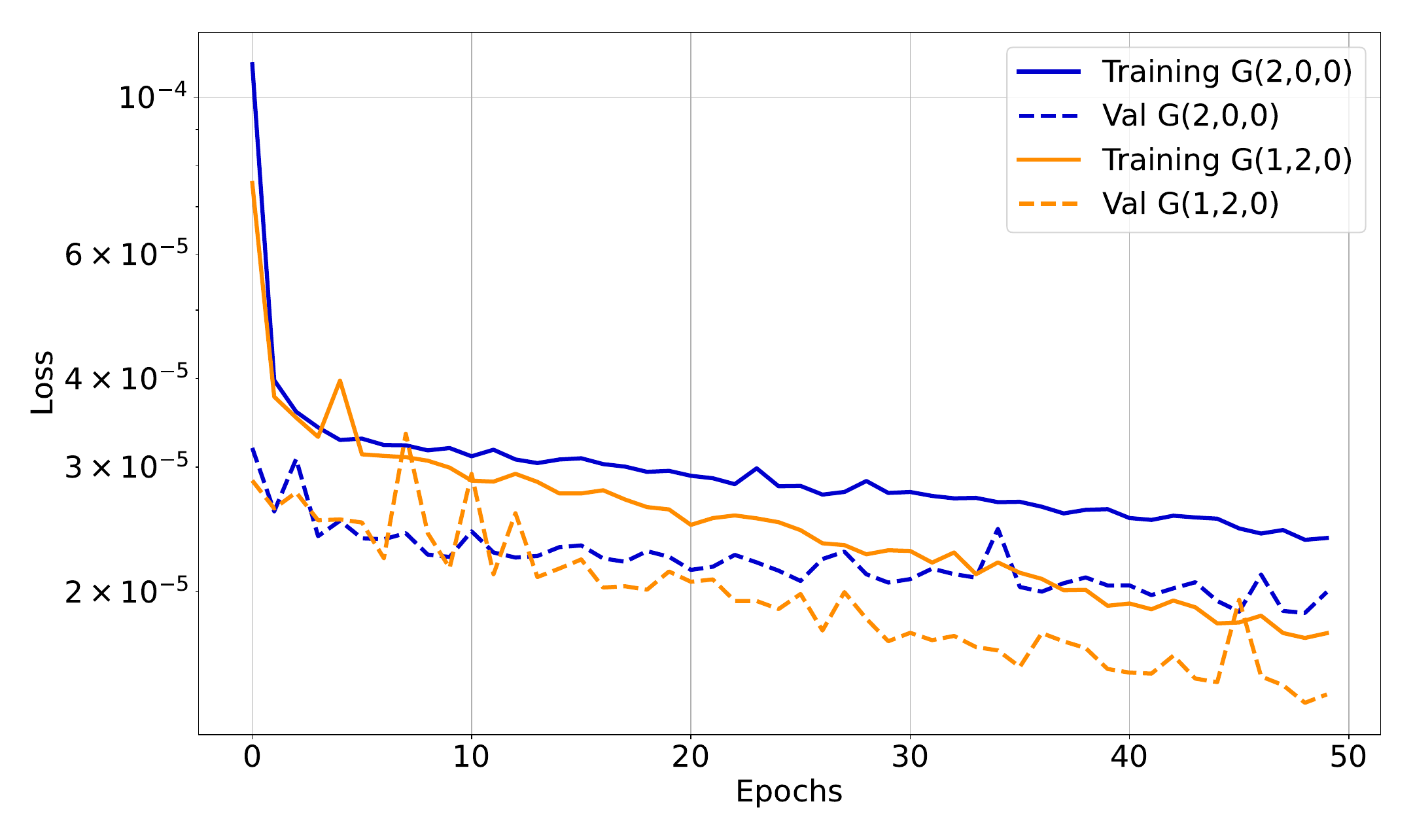}
            \caption[]%
            {{\small $\Delta t = 75$s}} 
            \label{fig:lossesdt75}
        \end{subfigure}
        \hfill
        \begin{subfigure}[b]{0.45\textwidth}   
            \centering 
            \includegraphics[width=\textwidth]{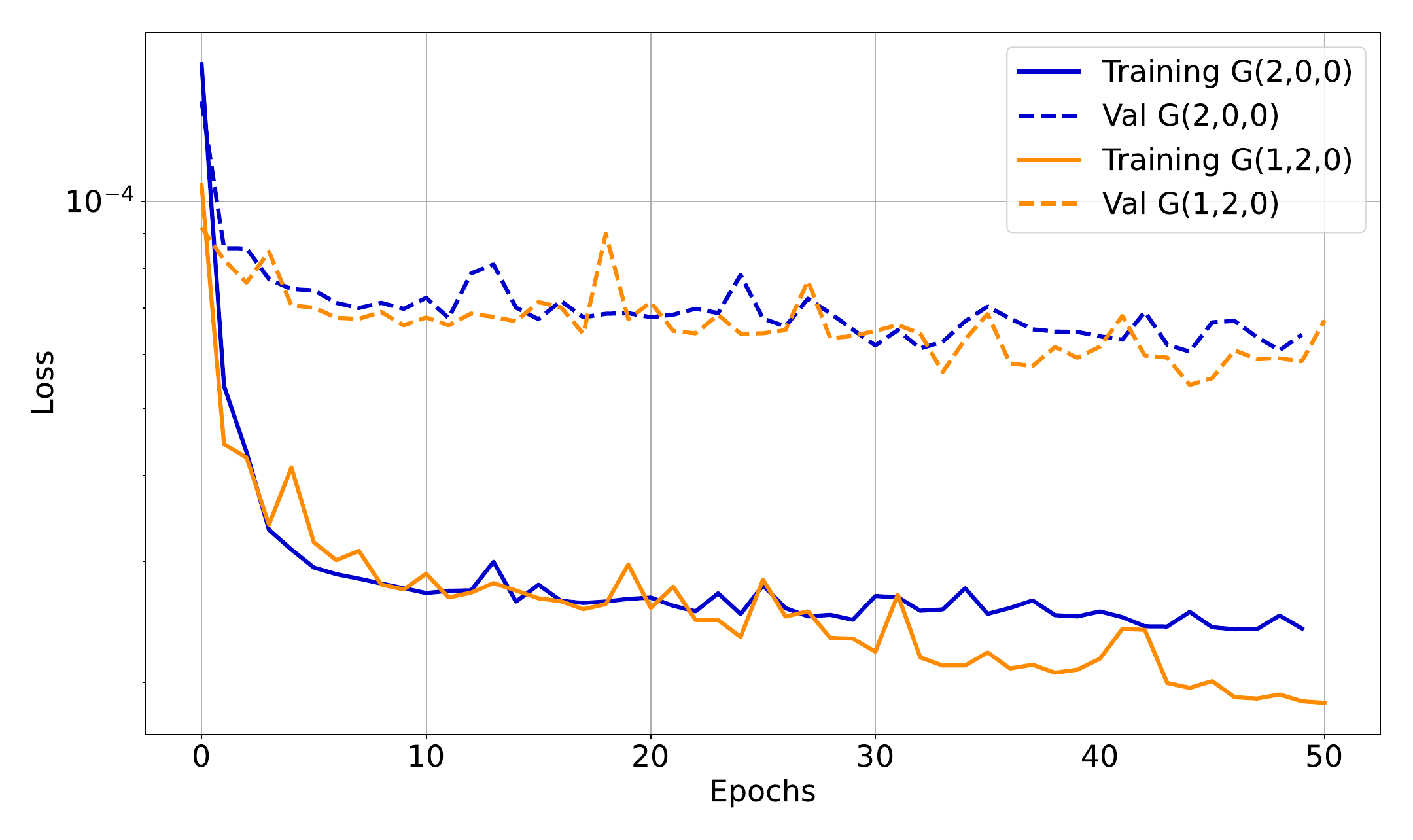}
            \caption[]%
            {{\small $\Delta t = 100$s}}    
            \label{fig:lossesdt100}
        \end{subfigure}
        \caption[ ]
        {\small Training and validation losses versus number of epochs for 2D Maxwell's PDEs for instances sampled at (a) $25$s, (b) $50$s, (c) $75$s, (d) $100$s.} 
        \label{fig:losses}
    \end{figure*}
    
\subsection{Training details}

Both networks have been trained for 50 epochs with a batch size of 32 in 2D and of 2 in the 3D case. We chose the Adam optimiser with learning rate of $10^{-3}$. The objective function to be minimised is the mean squared error (MSE) between ground truth fields $\mathbf{E}_{i +2\Delta t},\mathbf{B}_{i +2\Delta t}$ and estimated fields $\hat{\mathbf{E}}_{i +2\Delta t},\hat{\mathbf{B}}_{i +2\Delta t}$, which in 2D is defined as \begin{equation}
    \mathcal{L}_{2D} = \frac{1}{MN} \sum_{m=0}^M \sum_{n=0}^N  (E_{xmn,i +2\Delta t} - \hat{E}_{xmn,i +2\Delta t})^2 +
    (E_{ymn,i +2\Delta t} - \hat{E}_{ymn,i +2\Delta t})^2 + (B_{zmn,i +2\Delta t} - \hat{B}_{zmn,i +2\Delta t})^2,
    \label{loss}
\end{equation} in which $m,n$ indicate the spatial location of the field within the 2D surface. In the 3D case, we have
\begin{equation}
    \mathcal{L}_{3D} = \frac{1}{LMN} \sum_{j} \sum_{l=0}^L \sum_{m=0}^M \sum_{n=0}^N  (E_{jlmn,i +2\Delta t} - \hat{E}_{jlmn,i +2\Delta t})^2
    + (B_{jlmn,i +2\Delta t} - \hat{B}_{jlmn,i +2\Delta t})^2,
    \label{loss3D}
\end{equation} in which $l,m,n$ indicate the spatial location of the field within the 3D volume and $j = \{x,y,z\}$. Note how this is equivalent to measure the loss over the real coefficients of ground truth and estimated $F, \hat{F}$, for the vanilla GA approaches, and the real coefficients of ground truth and estimated $\mathbf{F}, \hat{\mathbf{F}}$, for the STA approach.
We compare results obtained with the weights of the two pipelines yielding minimum MSE error at validation stage.  The code has been implemented via tensorflow 2.13.1 and accelerated with CUDA 12.2 on an NVIDIA GeForce RTX 4090 GPU. Code can be found \href{https://github.com/albertomariapepe/STAResNet/tree/main}{here}.

\begin{figure}[!htpb]
        \centering
        \begin{subfigure}[b]{0.475\textwidth}
            \centering
            \includegraphics[width=\textwidth]{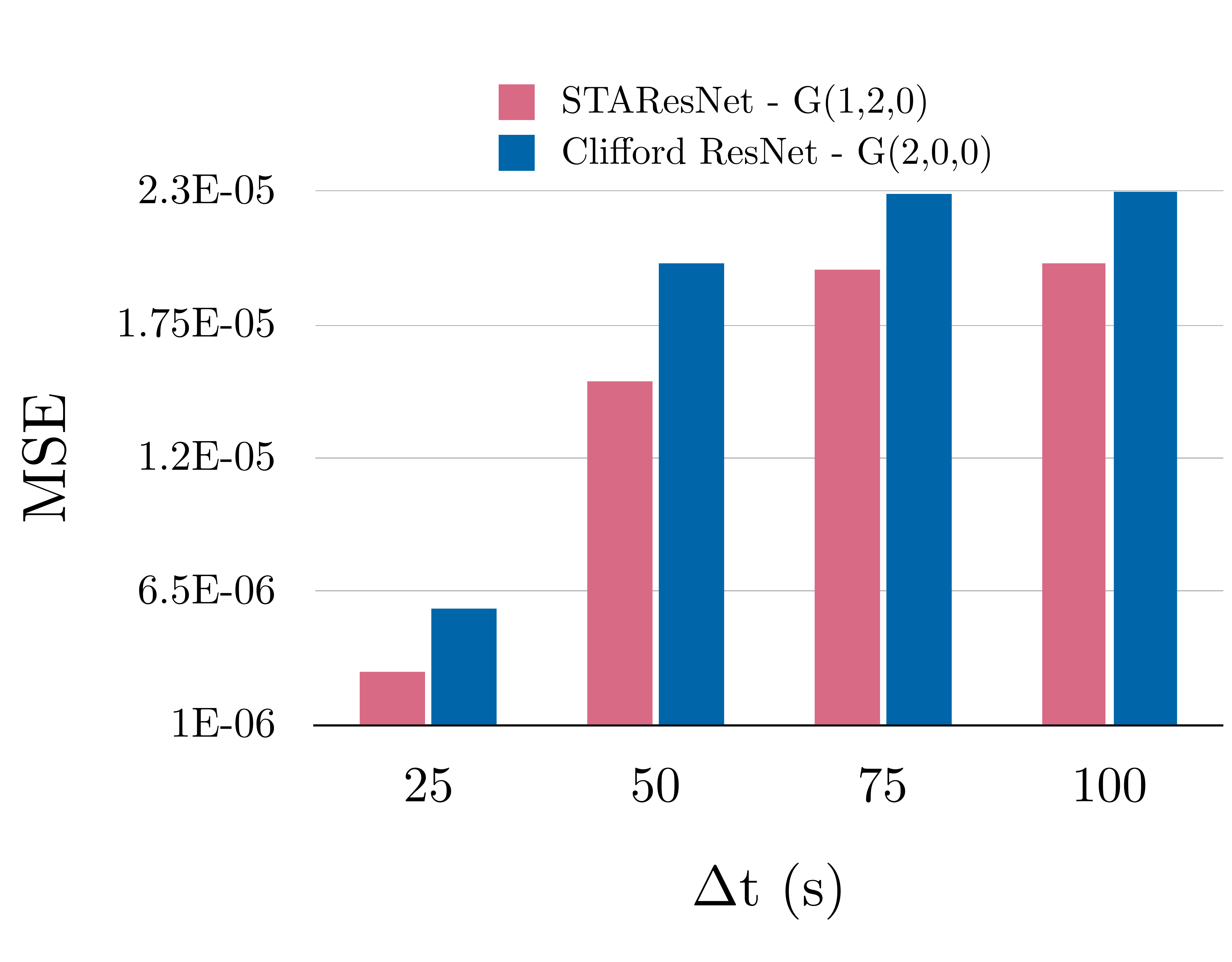}
            \caption[Network2]%
            {\small MSE ($\downarrow$) versus $\Delta t$.}    
            \label{fig:msevsdt}
        \end{subfigure}
        \hfill
        \begin{subfigure}[b]{0.475\textwidth}  
            \centering 
            \includegraphics[width=\textwidth]{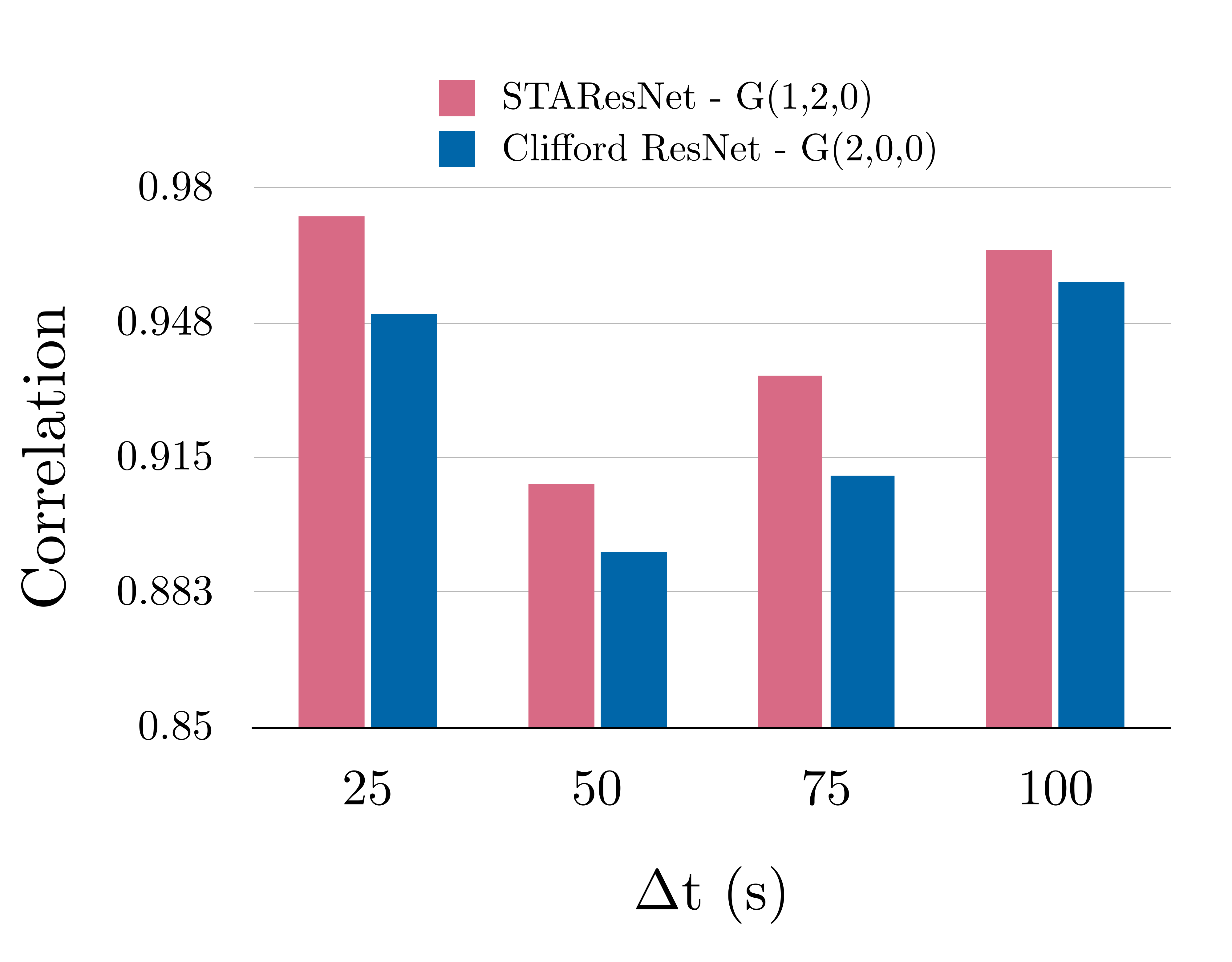}
            \caption[]%
            {\small Correlation ($\uparrow$) versus $\Delta t$.}    
            \label{fig:corrvsdt}
        \end{subfigure}
        \caption[ ]
            {\small (a) Mean squared error and (b) correlation between estimated and ground truth EM fields in the test set for varying $\Delta t$.} 

        \label{fig:mse}
    \end{figure}
    
\section{Experiments in 2D}

\subsection{Impact of sampling period}

The first scenario we look at is the solution of the Maxwell PDEs for varying sampling time $\Delta t$ of the EM field. The datasets have been generated with a Finite-Difference Time-Domain (FDTD) solver following the specifications of \cite{brandstetter2022clifford}. We consider a surface with spatial resolution of $32 \times 32$  and step size $\Delta x = \Delta y = 5\cdot 10^{-7}$m, with the EM field sampled at varying sampling period $\Delta t = \{25, 50, 75, 100\}$s. The light is propagated from 6 point sources randomly placed in the $xy$ plane.  The wavelength of the emitted light is $\lambda = 10^{-5}m$. Each light source emits light with a random phase and random amplitude. 

The training set includes 30000 frames, divided into 32 sequences of 12 samples of the EM field. For each sequence we employ the first two samples, i.e. the first two time steps, as input to the networks and the third as target. The validation and test sets are structured similary, and they include 3000 and 2400 frames, respectively. 

The training and validation loss profiles for the four $\Delta t$ cases are shown in Fig. \ref{fig:losses}. STAResNet consistently achieves lower loss both at both training and validation stages compared to its 2D GA counterpart.

\begin{figure}[!htbp]
        \centering
        \begin{subfigure}[b]{0.475\textwidth}
            \centering
            \includegraphics[width=\textwidth]{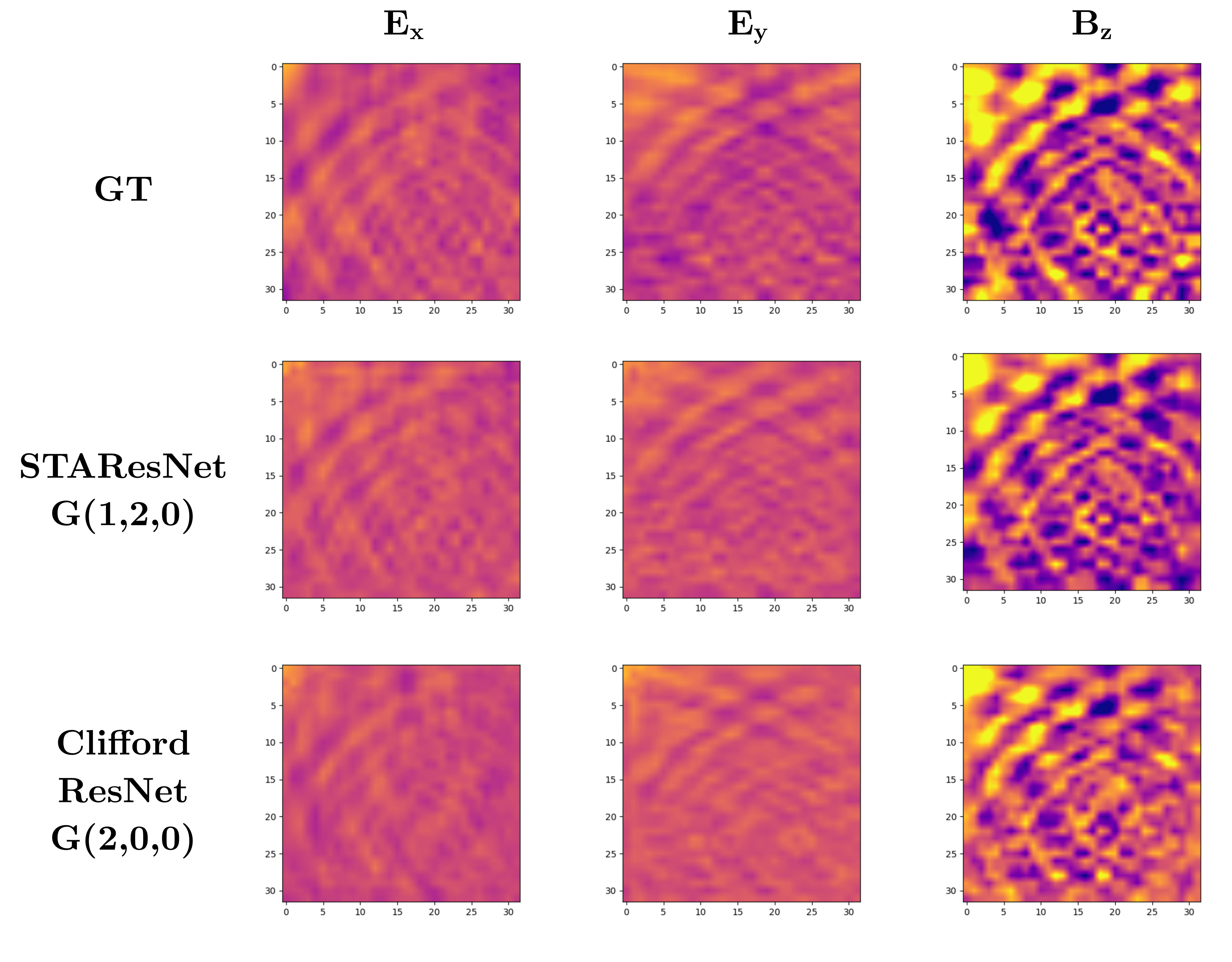}
            \caption[Network2]%
            {\small  $\Delta t = 25$s.}    
            \label{fig:dt25s}
        \end{subfigure}
        \hfill
        \begin{subfigure}[b]{0.475\textwidth}  
            \centering 
            \includegraphics[width=\textwidth]{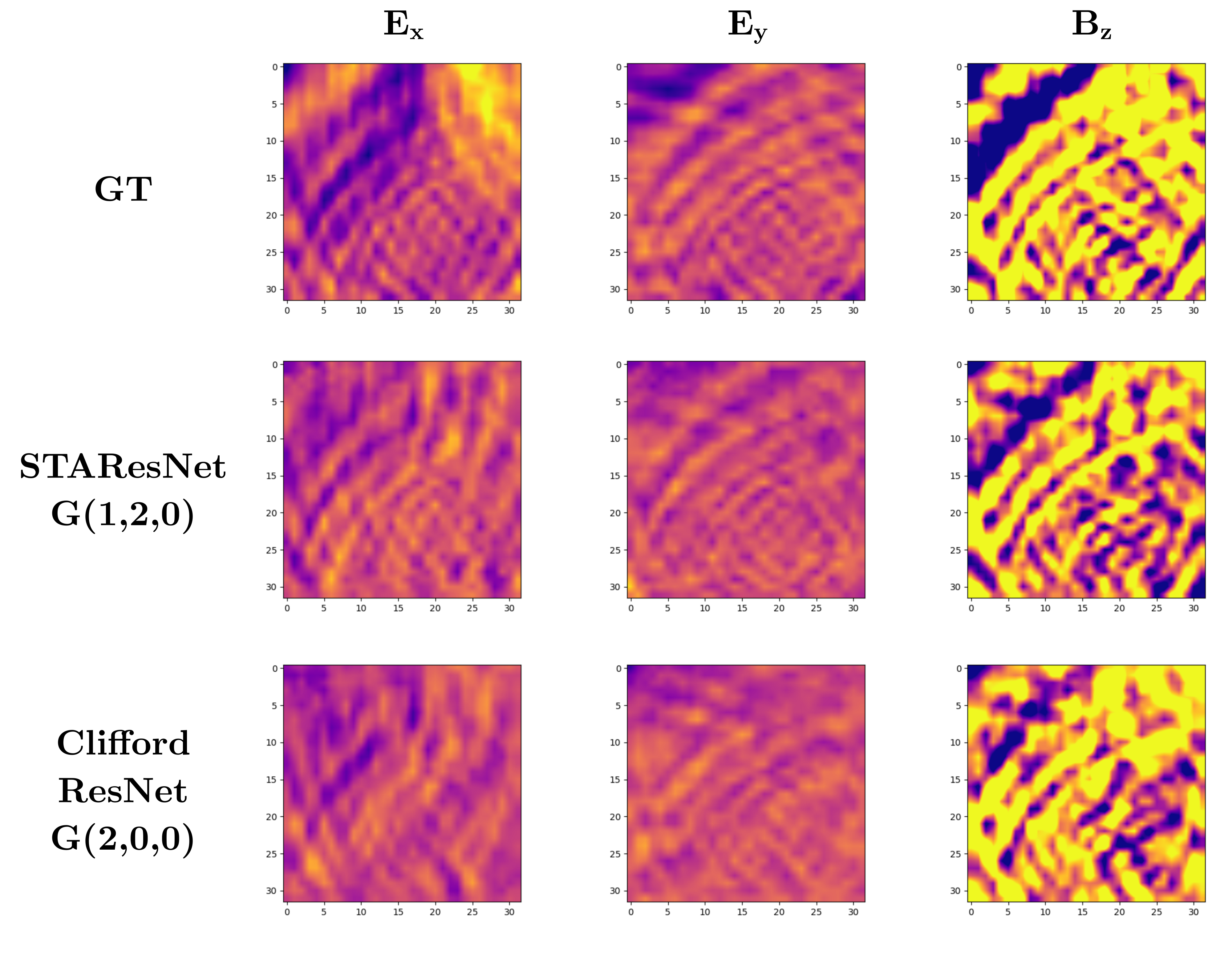}
            \caption[]%
            {\small $\Delta t = 50$s.}    
            \label{fig:dt50s}
        \end{subfigure}
        \caption[ ]
            {\small Ground truth and estimated EM field with Clifford ResNet and STAResNet for datasets sampled at (a) $\Delta t = 25$s and (b) $\Delta t = 50$s.} 

        \label{fig:ex11}
    \end{figure}

We measure two quantities at testing stage, the MSE as defined in Eq. \ref{loss} and the correlation index, defined in 2D as \begin{equation}
    r_{2D} = \frac{1}{MN}\sum_{m=0}^M \sum_{n=0}^N T_{mn} \hat{T}_{mn}
\end{equation}where $T = [E_{x,i+2\Delta t}, E_{y,i+2\Delta t}, B_{z,i+2\Delta t}]$ is the ground truth EM fields, $\hat{T} = [\hat{E}_{x,i+2\Delta t}, \hat{E}_{y,i+2\Delta t}, \hat{B}_{z,i+2\Delta t}]$ the estimated EM fields and $T_{mn}$ indicates the fields evaluated in position $(m,n)$ on the 2D grid. Results are summarised in Fig. \ref{fig:mse}. As expected, the MSE increases as more time passes between successive frames. Correlation increases as for large $\Delta t$ the fields tend to dissipate, hence presenting less busy patterns that result in higher correlation despite a worse estimation of the fields. STAResNet consistently performs better regardless of the sampling period $\Delta t$. 

\begin{figure}[!htbp]
    \centering
    \includegraphics[width=0.7\textwidth]{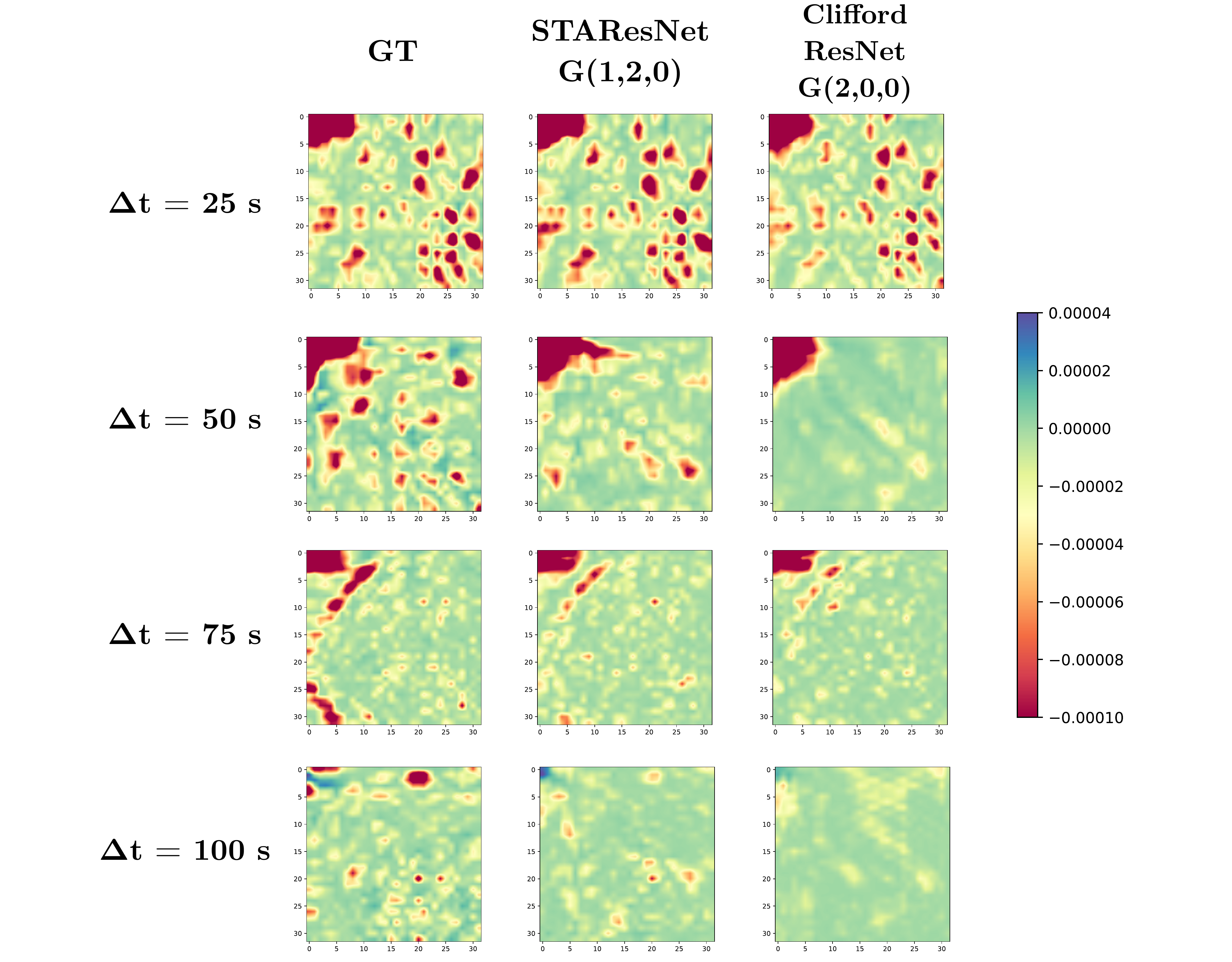}
    \caption{Squared magnitude of the Faraday bivector $\mathbf{F}^2$ for varying $\Delta t$.}
    \label{fig:faraday}
\end{figure}

Examples of the estimated electric and magnetic fields for varying $\Delta t$ are given in Fig. \ref{fig:ex11}. In all the cases presented, STAResNet does a better job at estimating the interference patterns arising from multiple sources as opposed to the Clifford ResNet. 
Additionally, STA has the advantage of unifying the fields components into a single object, the Faraday bivector. We can also choose to visualise the magnitude of the Faraday bivector defined as \begin{equation}
    \mathbf{F}^2 = (E_x \gamma_{10} + E_y \gamma_{20} + B_z \gamma_{12})^2
\end{equation} Note how, due to the STA signature, $\mathbf{F}^2$ can be negative. Examples of $\mathbf{F}^2$ plots for each tested $\Delta t$ are shown in Fig. \ref{fig:faraday} As expected, as $\Delta t$ increases the PDE solutions become less accurate, but in each of the four cases STAResNet yields magnitude plots with  more similar patterns to GT as opposed to Clifford ResNet working in 2D GA.

\subsection{Impact of obstacles}

We make the PDE solution more challenging by increasing the surface size to $48 \times 48$, and adding a single obstacle of varying dimension and fixed permittivity $\epsilon = 1.7^2$. The step size has been kept to $\Delta x = \Delta y = 5\cdot 10^{-7}$m and the sampling period to $\Delta t = 25$s. Light sources follow the same specifications of subsection 4.1.

\begin{figure}[!htbp]
    \centering
    \includegraphics[width=\textwidth]{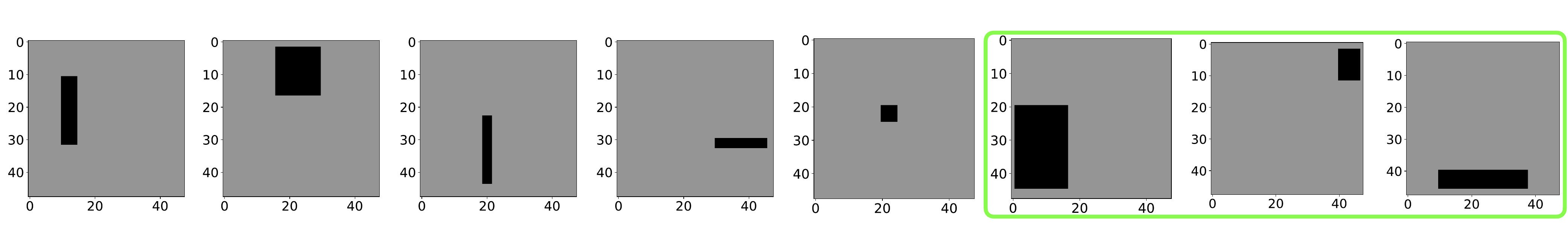}
    \caption{The 5 different obstacle configurations. The 3 unseen geometries are highlighted.}
    \label{fig:obst}
\end{figure}

The training set includes 30000 frames, divided into 32 sequences of 12 samples of the EM field, with five different obstacle configurations appearing with the same probability (i.e. 6000 frames per configuration). The validation includes 12800 sequences with the same five obstacle configurations (i.e. 2560 frames per configuration) but with different light sources with respect to the training set.  We employed two different test sets, with 12800 sequences each. The first test set presents obstacles in the same position of the training set but with different light sources. The second test set is generated with three obstacles in different locations to those in the training set, to assess the network's ability to generalise to previously unseen geometries (see Fig. \ref{fig:obst}). To make the comparison even more challenging, we reduced the number of channels of STAResNet from 24 to 23, for a total of 866326 trainable parameters against the 928580 parameters of Clifford ResNet.

\begin{figure}[!htbp]
    \centering
    \includegraphics[width=0.5\textwidth]{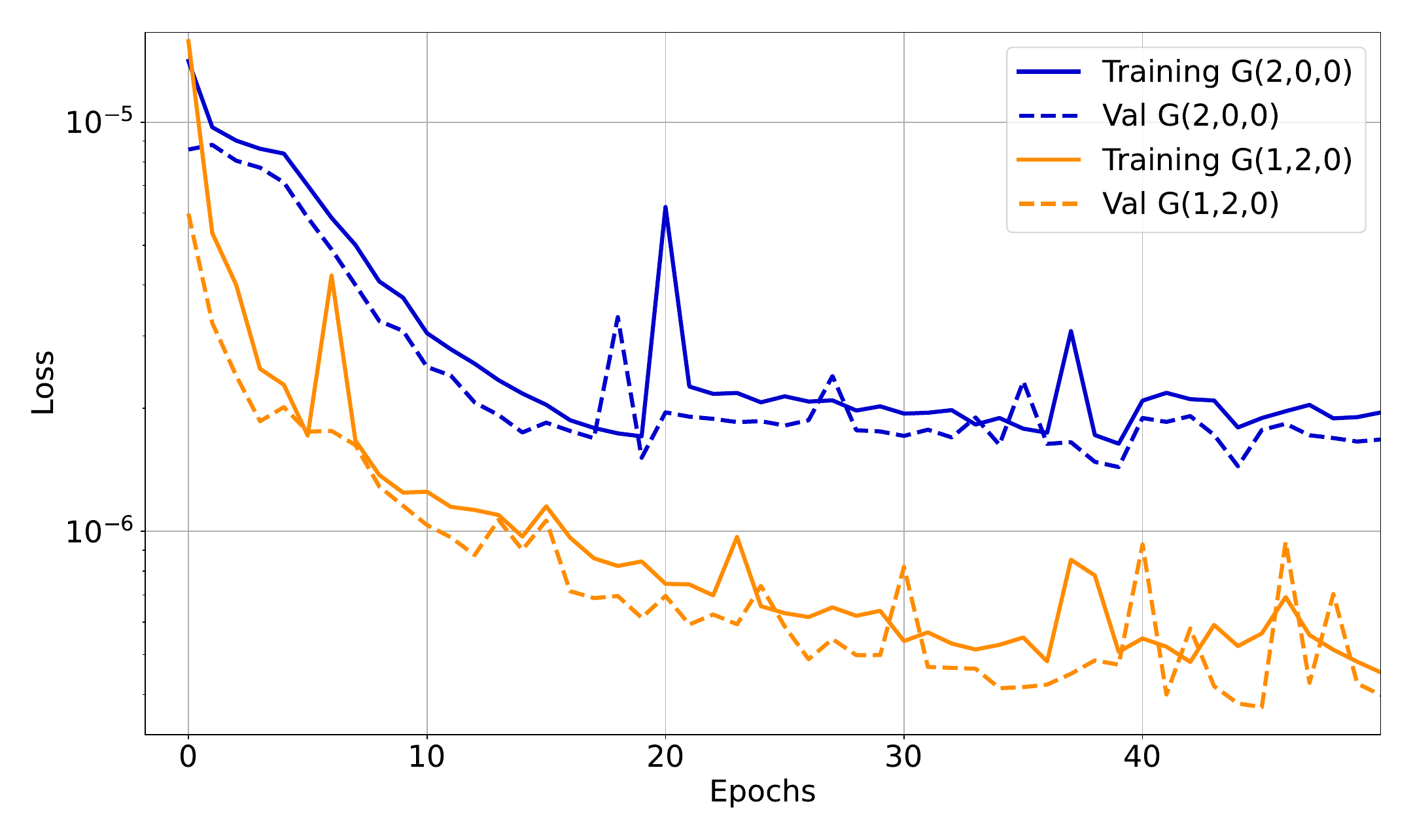}
    \caption{Training and validation losses versus number of epochs for the PDE in the presence of obstacles, for a total of 5 different obstacles configuration during training phase.}
    \label{fig:obstloss}
\end{figure}

\begin{figure}[!htbp]
        \centering
        \begin{subfigure}[b]{0.475\textwidth}
            \centering
            \includegraphics[width=\textwidth]{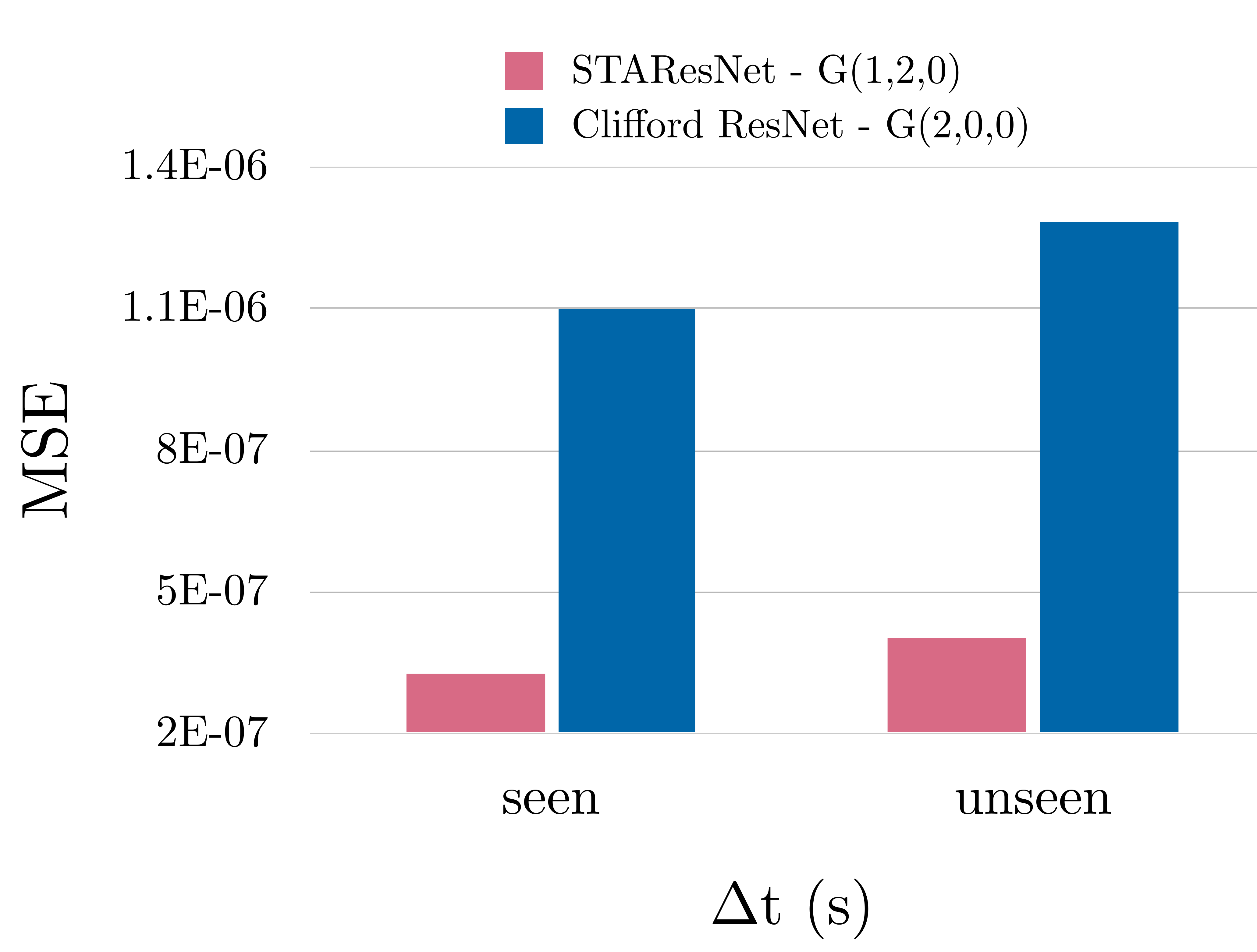}
            \caption[Network2]%
            {\small MSE ($\downarrow$).}    
            \label{fig:mseobst}
        \end{subfigure}
        \hfill
        \begin{subfigure}[b]{0.475\textwidth}  
            \centering 
            \includegraphics[width=\textwidth]{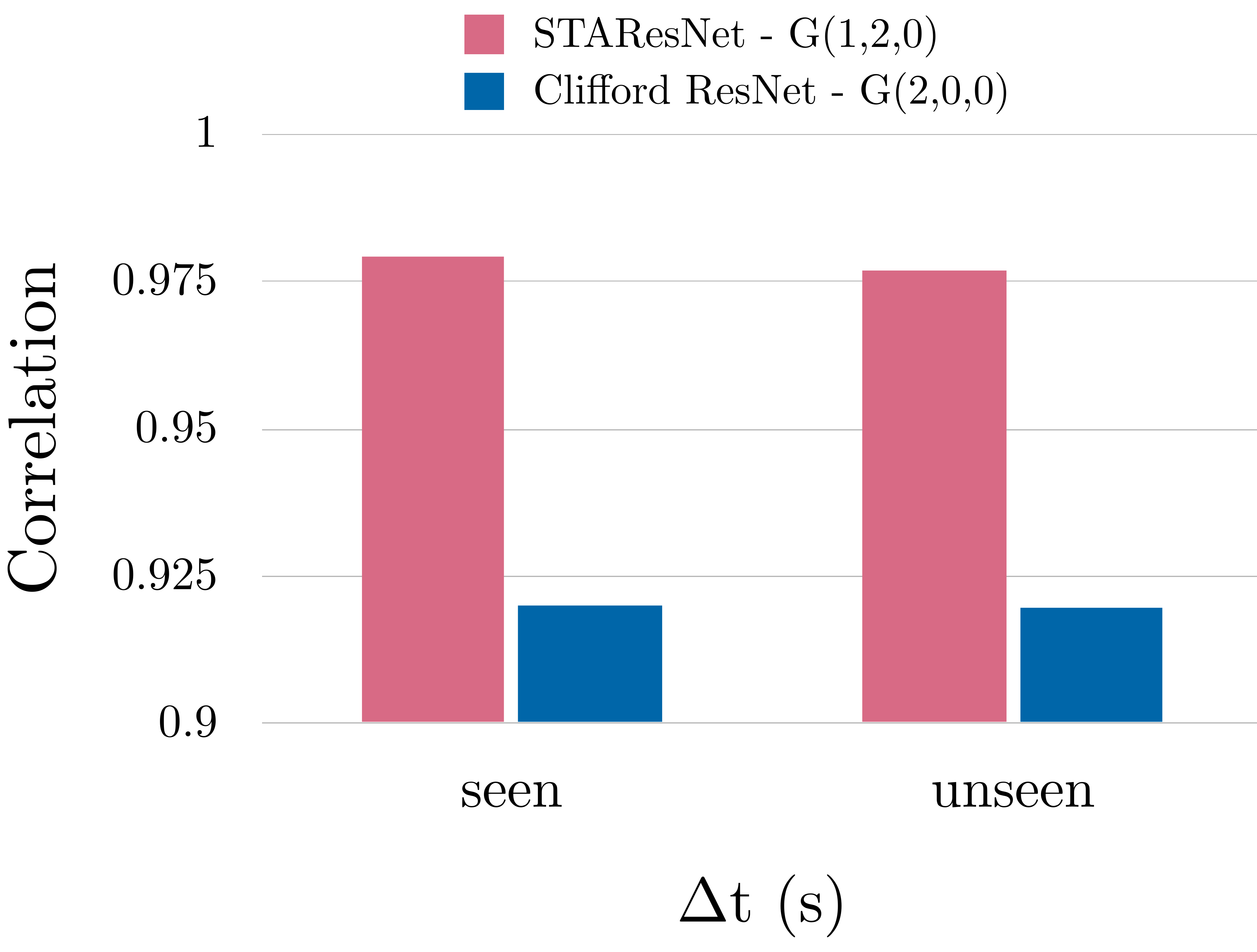}
            \caption[]%
            {\small Correlation ($\uparrow$).}    
            \label{fig:corrobst}
        \end{subfigure}
        \caption[ ]
            {\small (a) Mean squared error and (b) correlation between estimated and ground truth fields over the two test sets with seen and unseen obstacle configurations.} 

        \label{fig:mseobst2}
    \end{figure}

\begin{figure}[!htbp]
        \centering
        \begin{subfigure}[b]{0.475\textwidth}
            \centering
            \includegraphics[width=\textwidth]{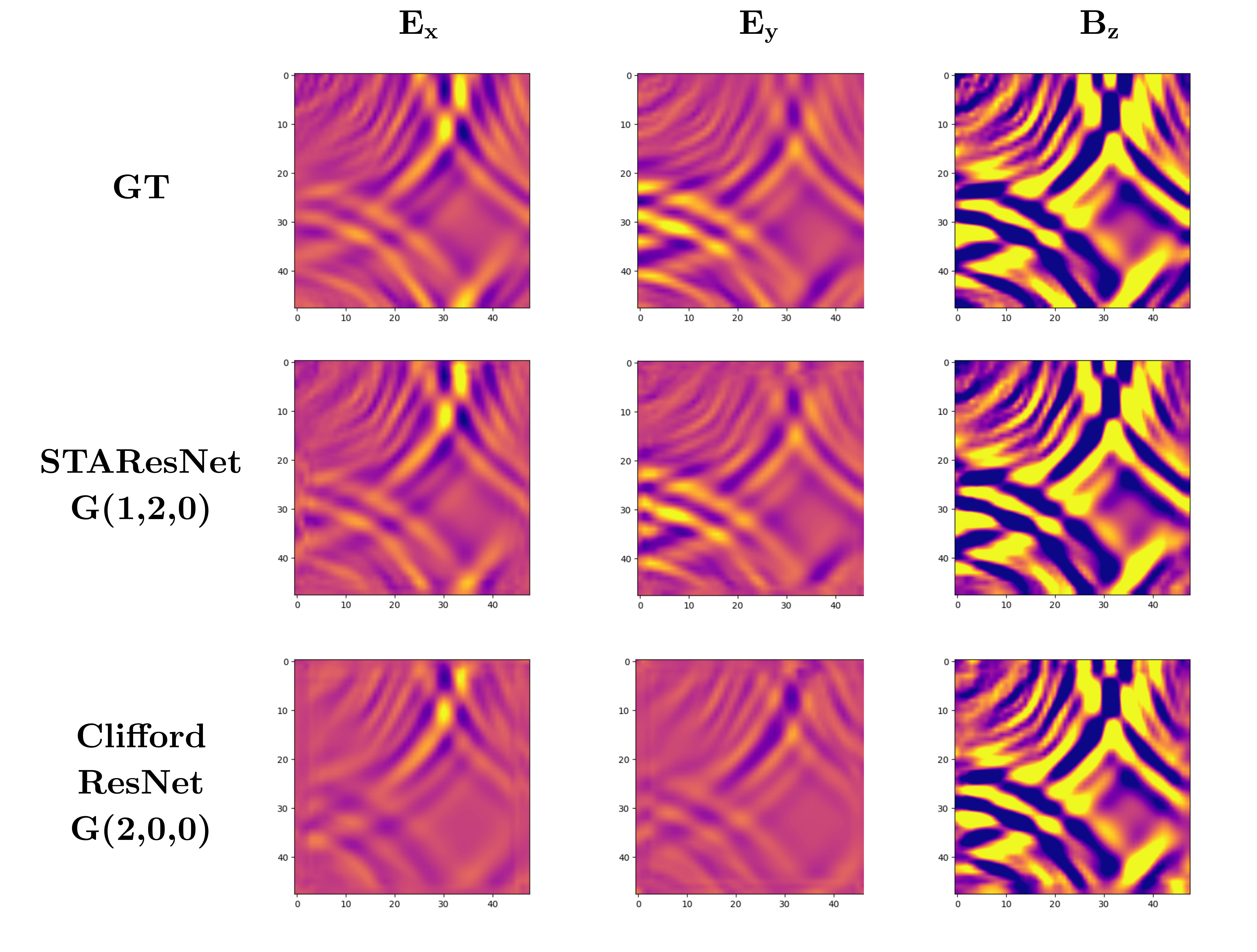}
            \caption[Network2]%
            {\small Estimated EM field for configuration 2.}    
            \label{fig:fieldsobst2}
        \end{subfigure}
        \hfill
        \begin{subfigure}[b]{0.475\textwidth}  
            \centering 
            \includegraphics[width=\textwidth]{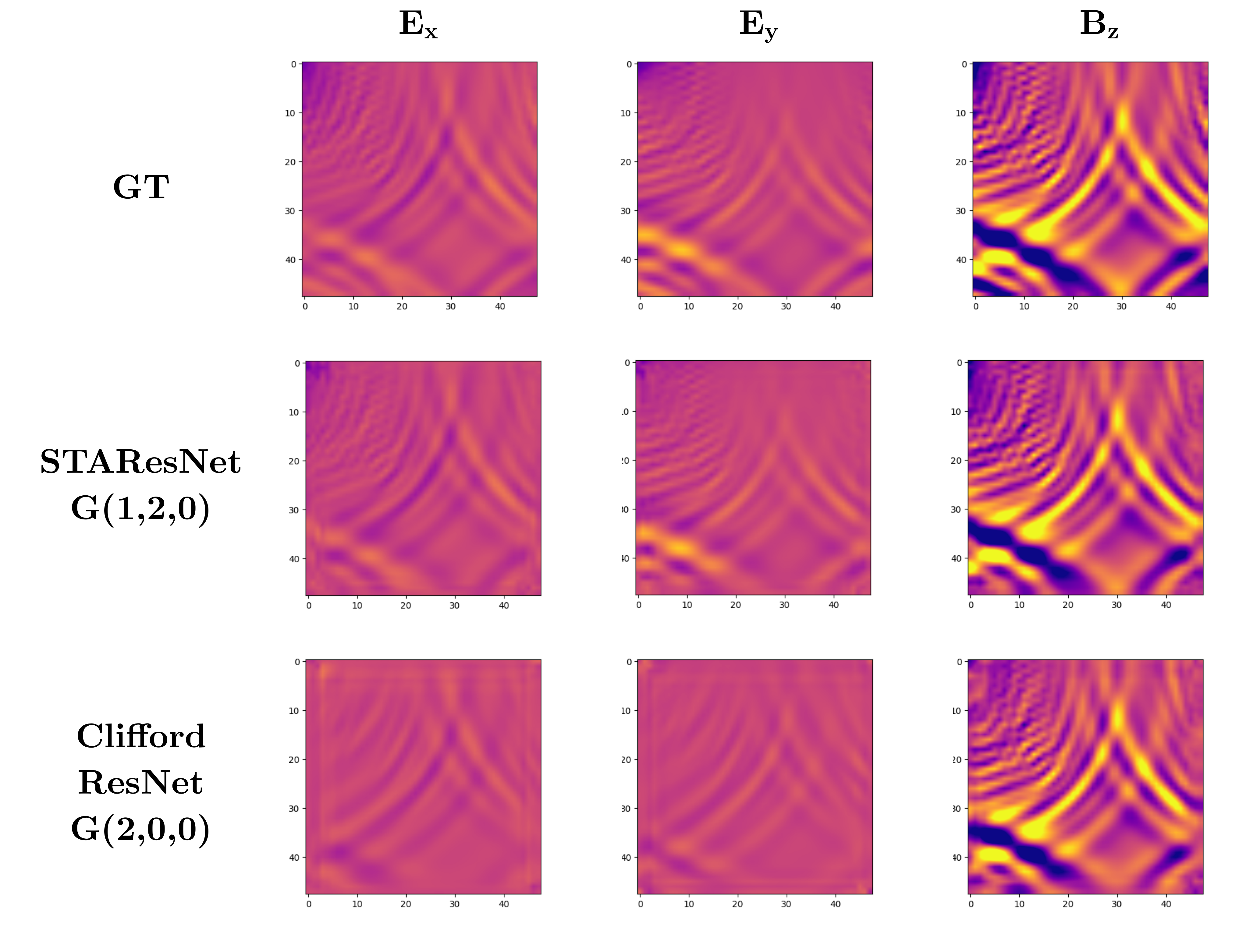}
            \caption[]%
            {\small Estimated EM field for configuration 5.}    
            \label{fig:fieldobst5}
        \end{subfigure}
        \caption[ ]
            {\small Ground truth and estimated EM field with Clifford ResNet and STAResNet for a field in the presence of obstacle in configurations (a) number 2 and (b) number 5 of Fig. 7.} 

        \label{fig:extobst}
    \end{figure}

Training and validation losses for Clifford ResNet and STAResNet are shown in Fig. \ref{fig:obstloss}. The trend is the same as that shown in Fig. \ref{fig:losses}: again, in the presence of obstacles with fixed sampling period, STAResNet attains a lower error. Moreover, STAResNet achieves lower MSE and higher correlation coefficient as compared to 2D Clifford ResNet regardless if testing on seen or unseen obstacle configuration (see Fig. \ref{fig:mseobst2}). 
This is clearly visible in Fig. \ref{fig:extobst}: estimated fields with STAResNet preserve more structure than those estimated with Clifford ResNet.

\begin{figure}[!htbp]
    \centering
    \includegraphics[width=0.7\textwidth]{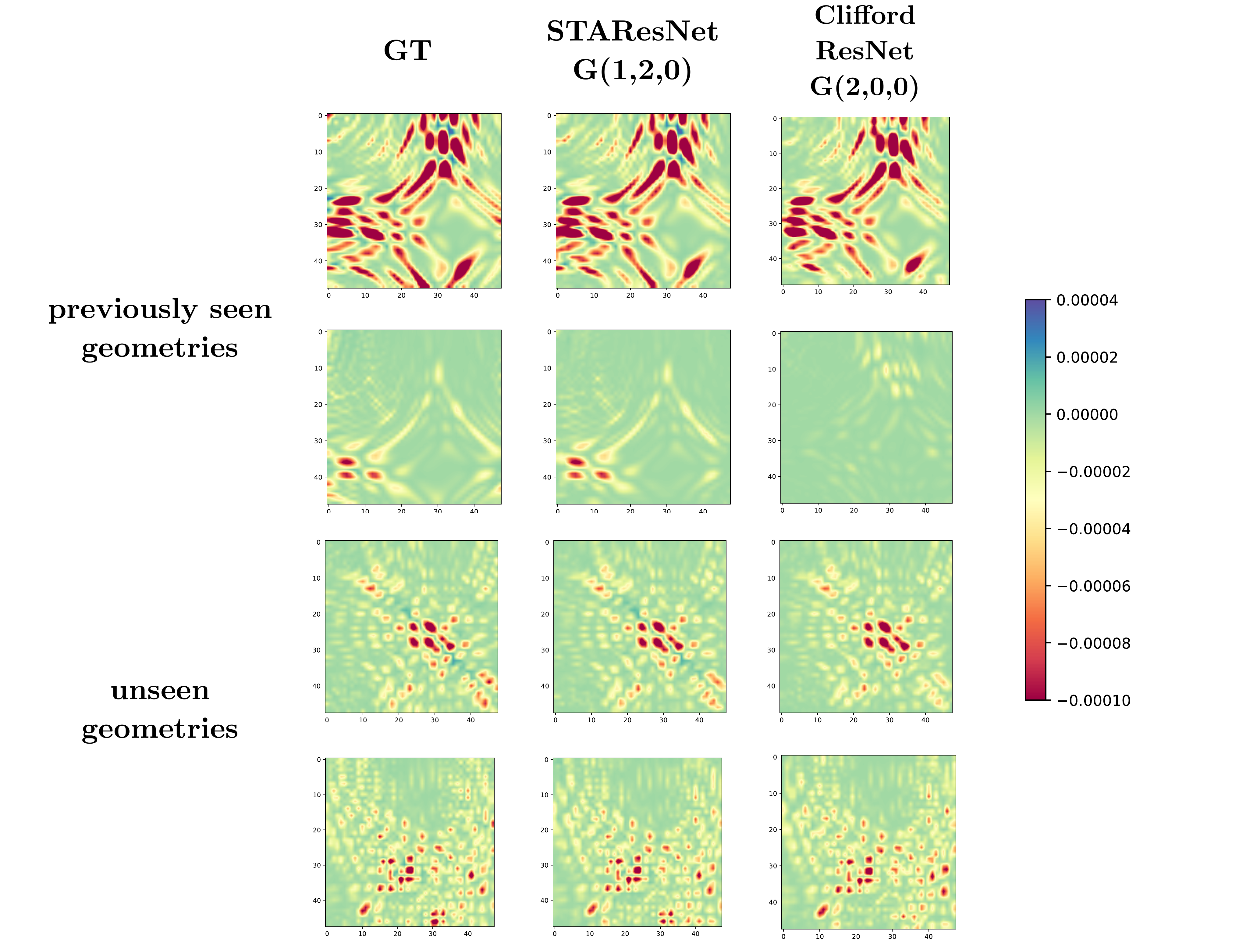}
    \caption{Squared magnitude of the Faraday bivector $\mathbf{F}^2$ over the test set with seen obstacles configurations (1 to 5) and unseen obstacles configurations (6 to 8) as shown in Fig. 7.}
    \label{fig:faraday_obst}
\end{figure}

Examples of the squared magnitude of $\mathbf{F}$ estimated in the presence of obstacles are given in Fig. \ref{fig:faraday_obst}: also in this scenario, STAResNet better preserves the finer details in the Faraday bivector as opposed to a vanilla GA approach.

\subsection{Impact of number of parameters}

STAResNet works in a mathematical space with one additional dimension compared to Clifford ResNet in 2D GA. This means that the tensor embedded in G(1,2,0) in STAResNet will have its last dimension of size $2^3 = 8$, as opposed to $2^2 = 4$ in 2D Clifford ResNet. This impacts the number of trainable parameters of the two pipelines, meaning that for the same number of channels or for the same size of the convolutional filters, STAResNet will generally have a larger number of trainable parameters.

\begin{figure}[!htbp]
    \centering
    \includegraphics[width=0.7\textwidth]{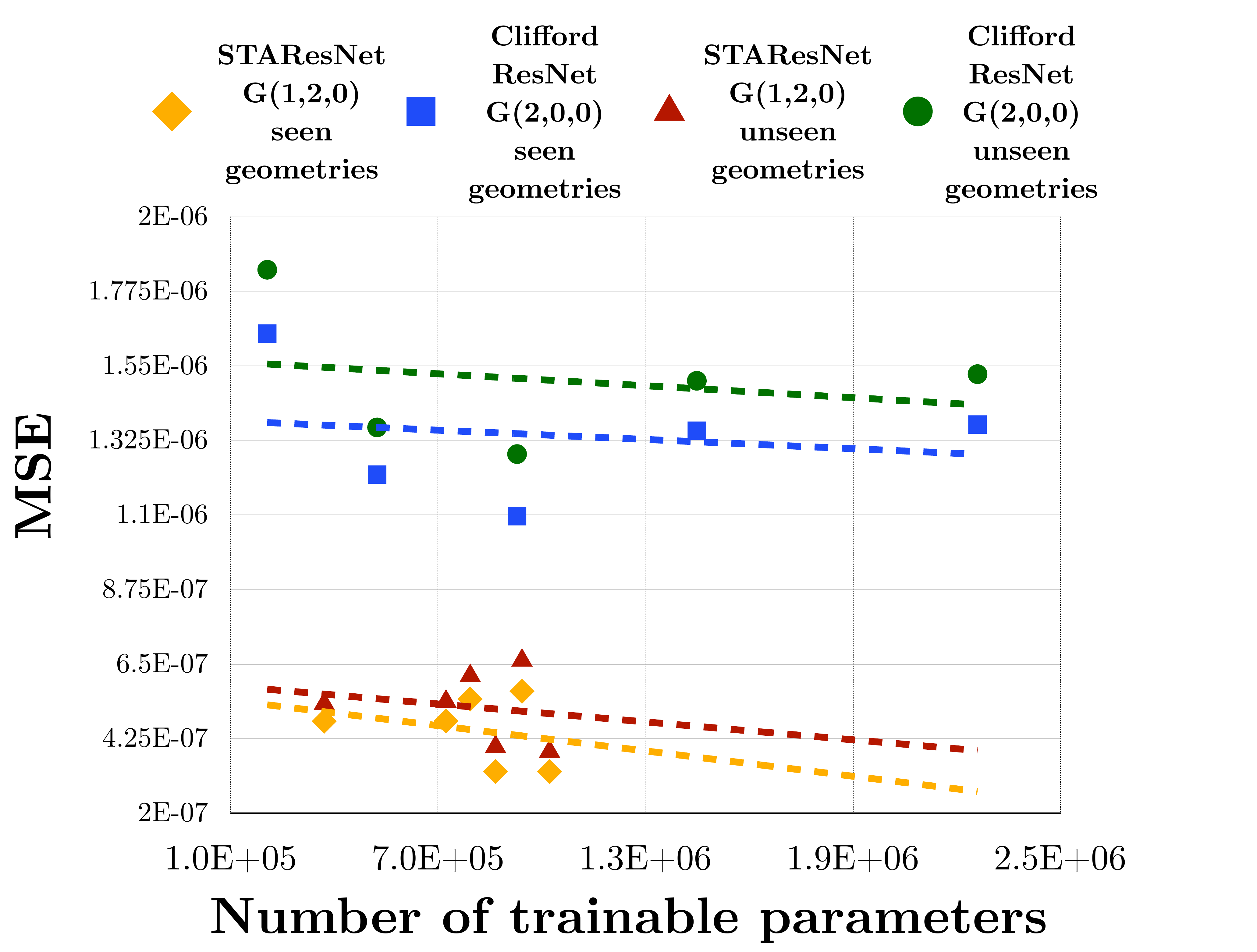}
    \caption{Test error over the estimated EM fields in the presence of seen and unseen obstacle geometries versus the number of trainable parameters.}
    \label{fig:trainparam}
\end{figure}

To verify that STAResNet is consistently superior to a 2D GA approach and provide a fair comparison, we train both networks with a varying number of hidden channels, ranging between 15 and 40. This yields pipelines with different number of trainable parameters, from $2 \times 10^5$ up to $2 \times 10^6$. Results are summarized in Fig. \ref{fig:trainparam}. 

The first thing it can be noticed is that the error decreases for a larger number of parameters in both networks, as expected. The test MSE error with 2D Clifford ResNet, however, never reaches values below $1 \times 10^{-6}$. Even with 40 channels and above 2 million trainable parameters, the 2D Clifford ResNet cannot catch up with STAResNet, that with just 15 channels and 370,000 parameters yields a test error of below $5.5 \times 10^{-7}$. This means that our 2D STAResNet, with 16.4$\%$ the number of parameters of 2D Clifford ResNet, can estimate EM fields twice as accurately. This proves that the improvement offered by STAResNet comes from the mathematical space it works in rather than the fact that it is generally a larger network. 

The second thing that can be seen in Fig. \ref{fig:trainparam} is the robustness to previously unseen data. The gap in the error between seen and unseen obstable configurations with STAResNet is visibly smaller than the gap with Clifford ResNet, which is less capable of generalisation when presented with test data which were not included in the training set. This is true regardless of the number of parameters of the networks.

\begin{figure}[!htbp]
        \centering
        \begin{subfigure}[b]{0.475\textwidth}
            \centering
            \includegraphics[width=\textwidth]{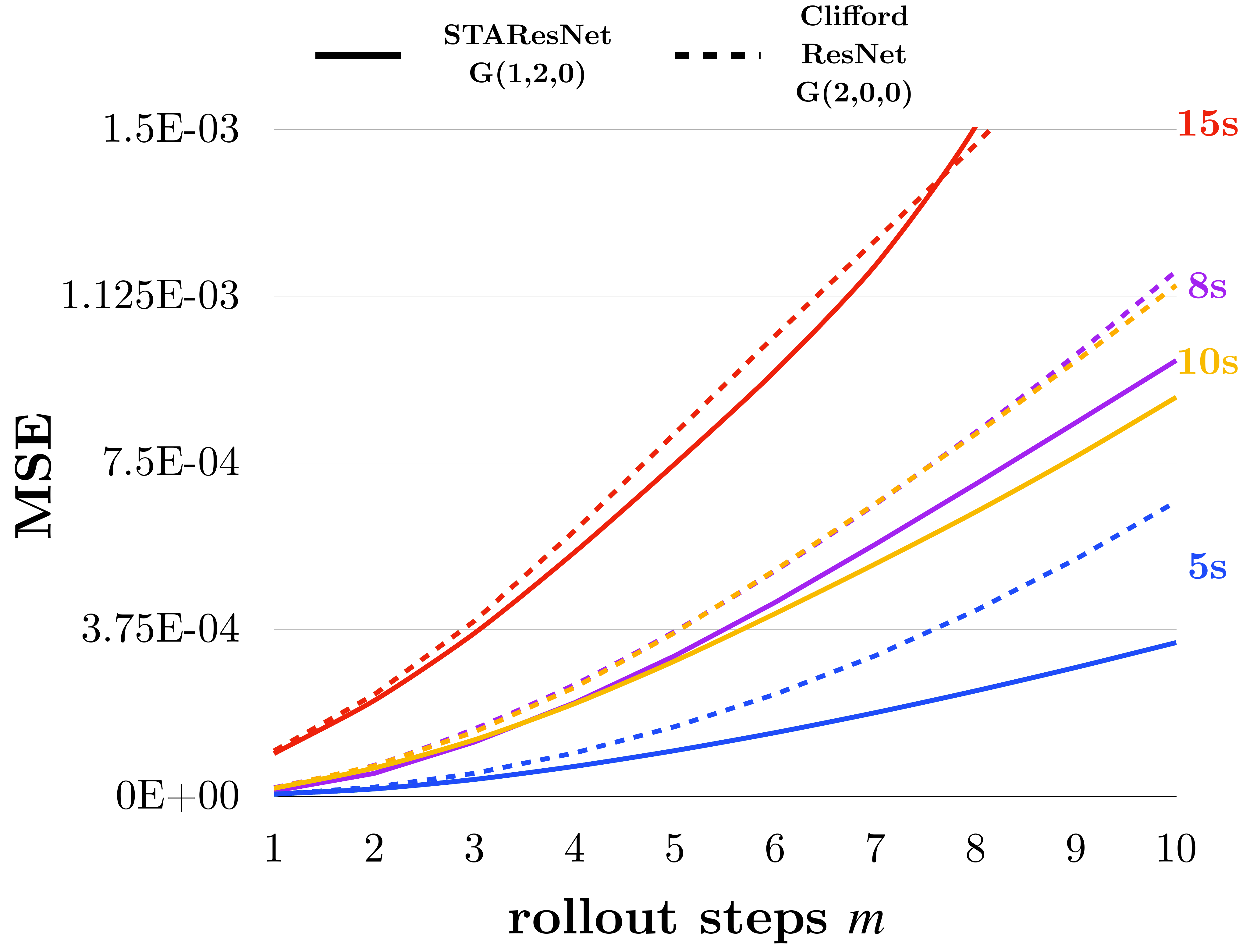}
            \caption[Network2]%
            {\small MSE ($\downarrow$) versus rollout steps.}    
            \label{fig:rolloutmse}
        \end{subfigure}
        \hfill
        \begin{subfigure}[b]{0.475\textwidth}  
            \centering 
            \includegraphics[width=\textwidth]{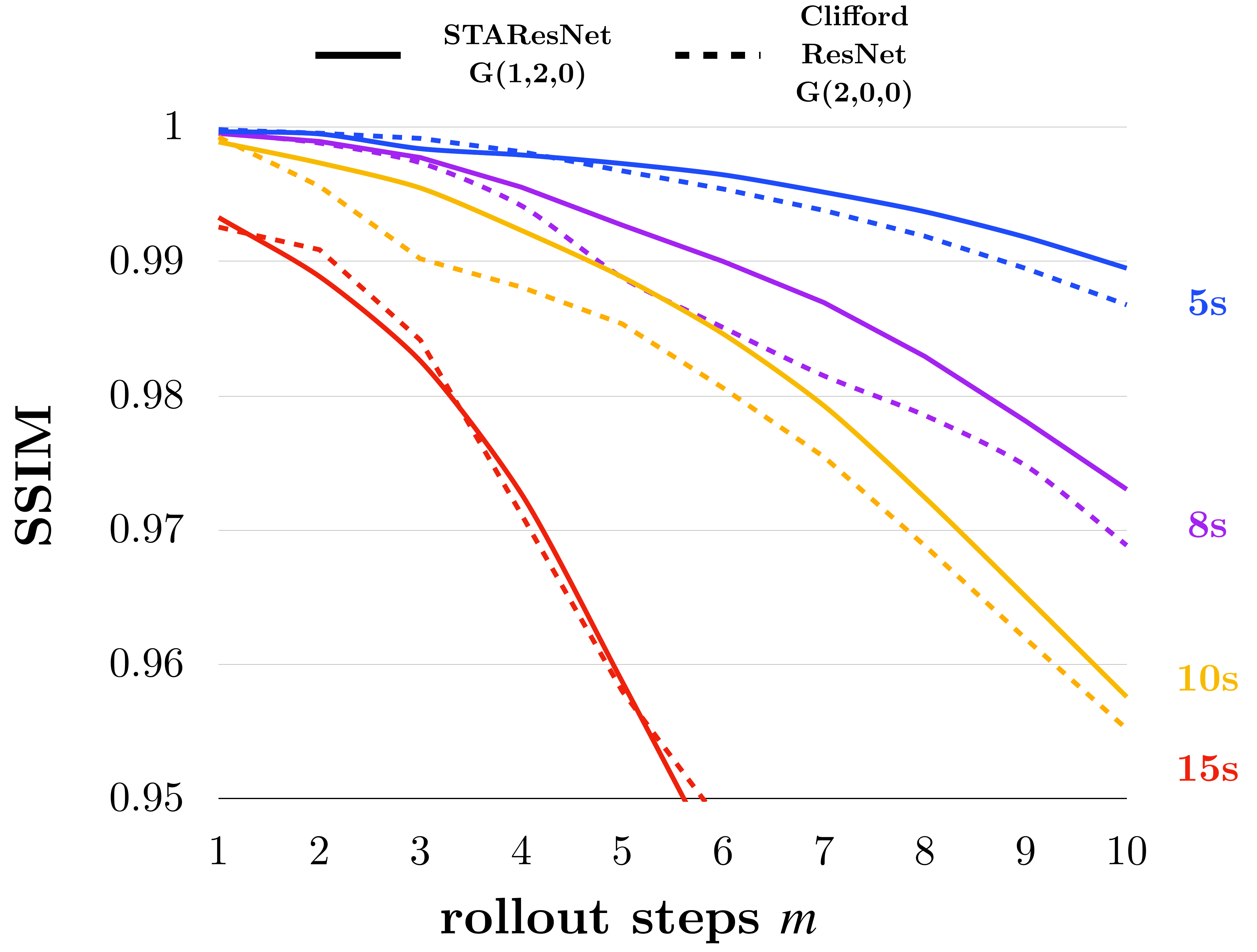}
            \caption[]%
            {\small SSIM ($\uparrow$) versus rollout steps.}    
            \label{fig:rolloutssim}
        \end{subfigure}
        \caption[ ]
            {\small (a) Mean squared error and (b) correlation between estimated and ground truth EM fields over test set versus rollout steps $m$ for the 2D case.} 

        \label{fig:rollout}
    \end{figure}
    
\subsection{Impact of rollout}

Lastly, we quantify the rollout error yielded by the two models. In the context of sequential modeling, \emph{rollout} refers to the process of using the model's own predictions as inputs to generate future predictions. For example, at rollout step $m=1$,   the model predicts \( y_{t + 2\Delta t} \) using \( y_t \) and \( y_{t + \Delta t} \), both ground truth. At the next rollout step, the model predicts \( y_{t + 3\Delta t} \) using \( y_{t \Delta t} \) and \( y_{t + 2\Delta t} \), in which now one of the inputs is the model's output at a previous time step. Successive time steps are estimated in a sliding window fashion, by feeding the model with its own outputs. It is desirable that the rollout error stays bounded in a model as it is much more likely that the model will be employed to estimate the PDE evolution in time rather than an isolated snapshot. 

Since each trajectory in our datasets contains 12 samples, we measure the rollout error for $m$ ranging between 1 and 10. Results are summarized in Fig. \ref{fig:rollout}, in which we measure MSE and structural similarity index (SSIM) between GT and predicted fields. Since the error rapidly propagates between successive time steps, we train Clifford ResNet and STAResNet on datasets captured for small $\Delta t$, i.e. $\Delta t = \{5, 8, 10, 15\}s$.

\begin{figure*}[!htbp]
        \centering
        \begin{subfigure}[b]{0.475\textwidth}
            \centering
            \includegraphics[width=\textwidth]{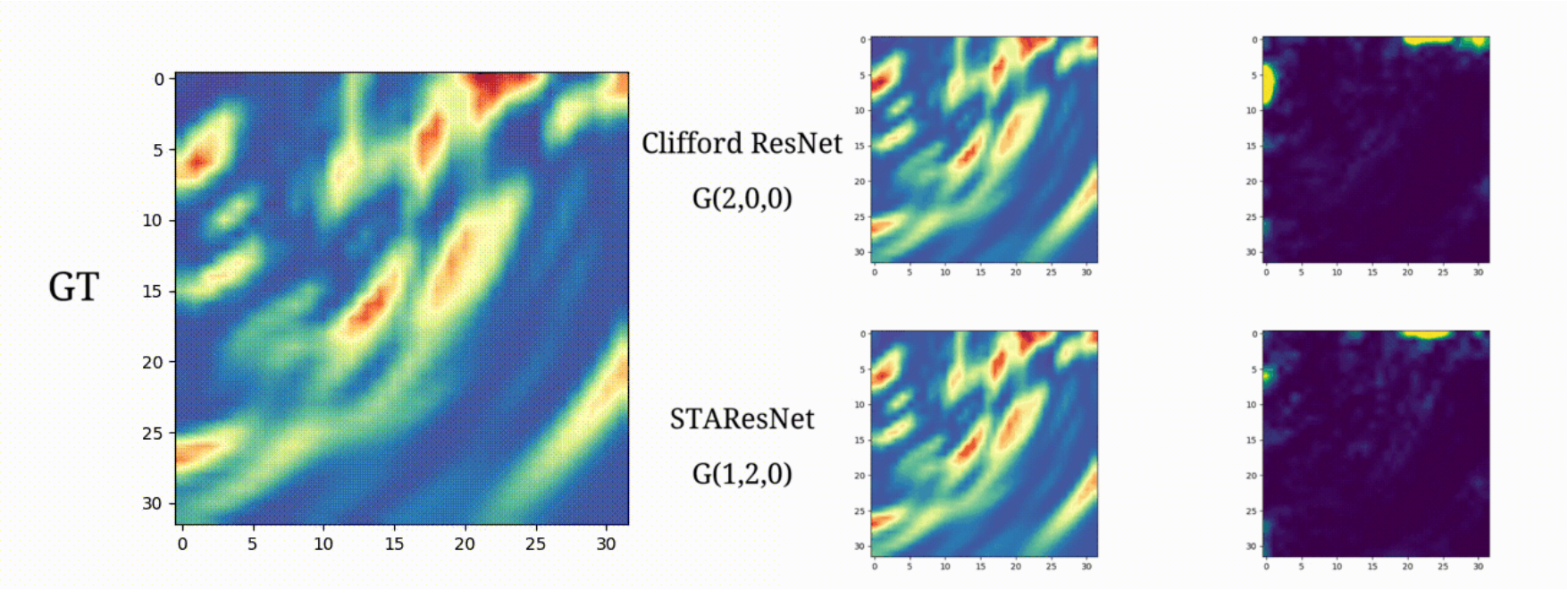}
            \caption[Network2]%
            {{\small $m$ = 1}}    
            \label{fig:meq1}
        \end{subfigure}
        \hfill
        \begin{subfigure}[b]{0.475\textwidth}  
            \centering 
            \includegraphics[width=\textwidth]{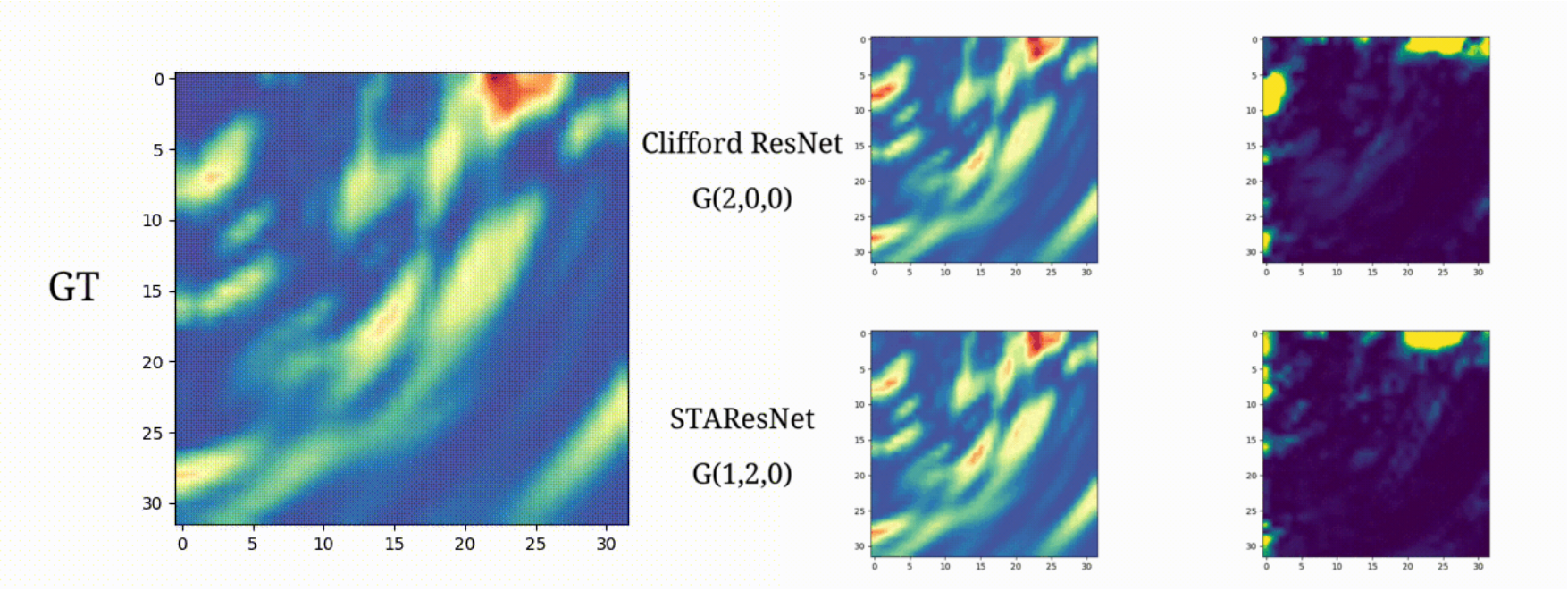}
            \caption[]%
            {{\small $m$ = 2}}    
            \label{fig:meq2}
        \end{subfigure}
        \vskip\baselineskip
        \begin{subfigure}[b]{0.475\textwidth}   
            \centering 
            \includegraphics[width=\textwidth]{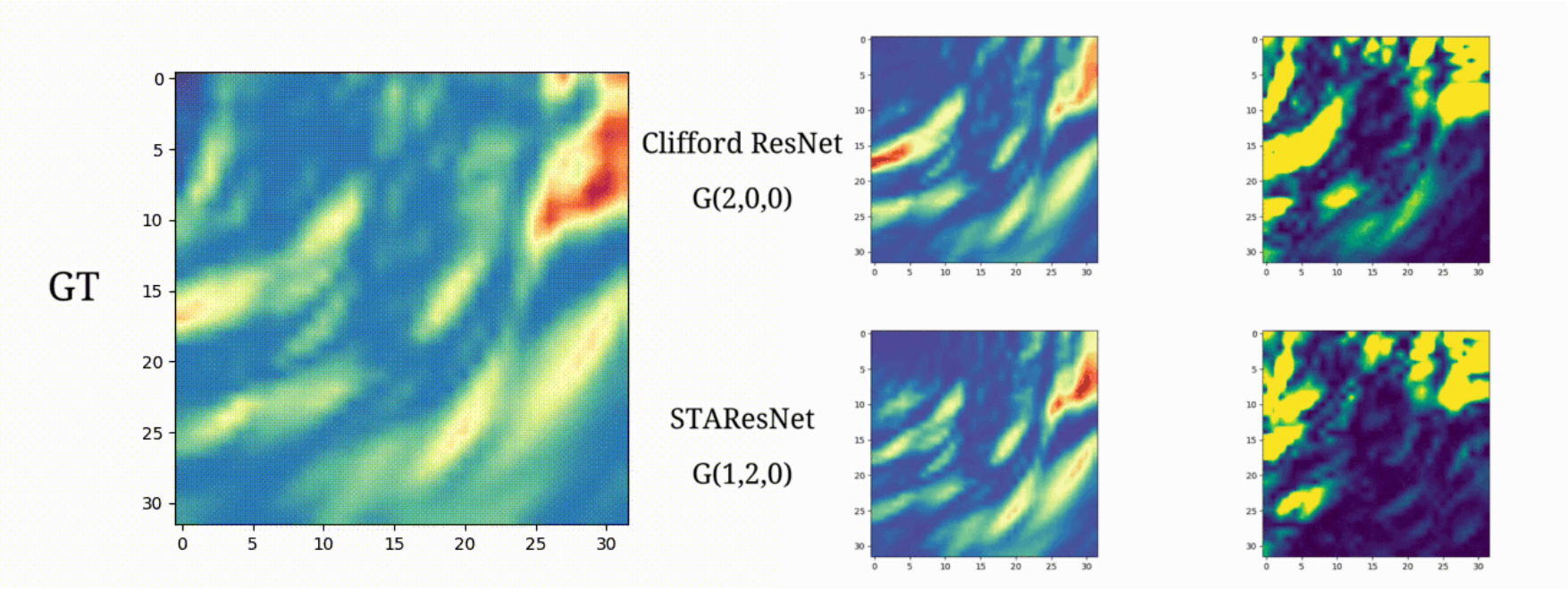}
            \caption[]%
            {{\small $m$ = 8}} 
            \label{fig:meq3}
        \end{subfigure}
        \hfill
        \begin{subfigure}[b]{0.475\textwidth}   
            \centering 
            \includegraphics[width=\textwidth]{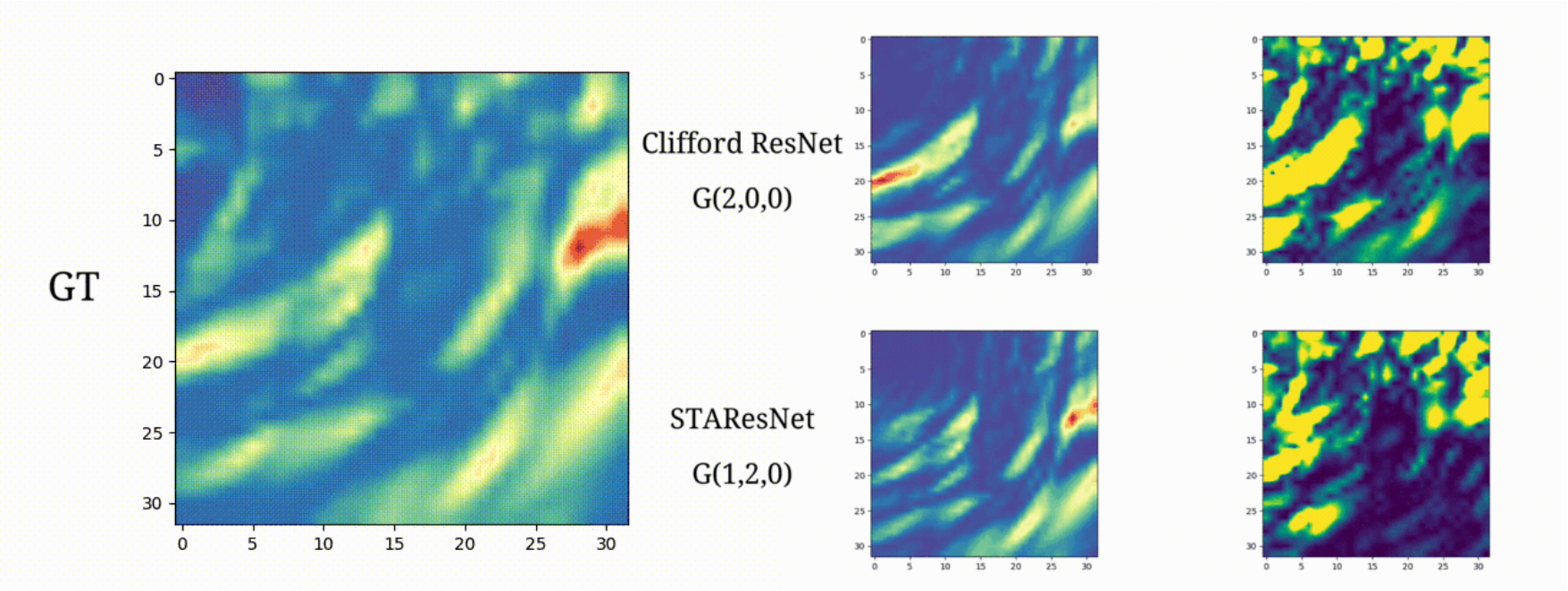}
            \caption[]%
            {{\small $m$ = 10}}    
            \label{fig:meq10}
        \end{subfigure}
        \caption[ ]
        {\small 2D GT $\mathbf{F}^2$, estimated $\hat{\mathbf{F}}^2$ and difference $|\mathbf{F}^2 - \hat{\mathbf{F}}^2|$ for Clifford ResNet and STAResNet at different rollout steps: (a) $m=1$, (b) $m=2$, (c) $m=8$ and (d) $m=10$. Here $\Delta t = 5$s.} 
        \label{fig:rolloutexamples}
    \end{figure*}

From Fig. \ref{fig:rollout} it is possible to notice that STAResNet approach yields consistently lower MSE and higher SSIMover Clifford ResNet, and that the gap between the two becomes more visible as the number of rollout steps increases. It can be argued that, as $\Delta t$ increases, the gap between the two models tends to narrow. This happens because for large $\Delta t$ there is little continuity between successive time steps, meaning that the error is already significant for small $m$, and for large $m$ both estimates with STAResNet and Clifford ResNet deviate so much with respect to GT to the comparison meaningless.

\begin{figure*}[!htpb]
        \centering
        \begin{subfigure}[b]{0.475\textwidth}
            \centering
            \includegraphics[width=\textwidth]{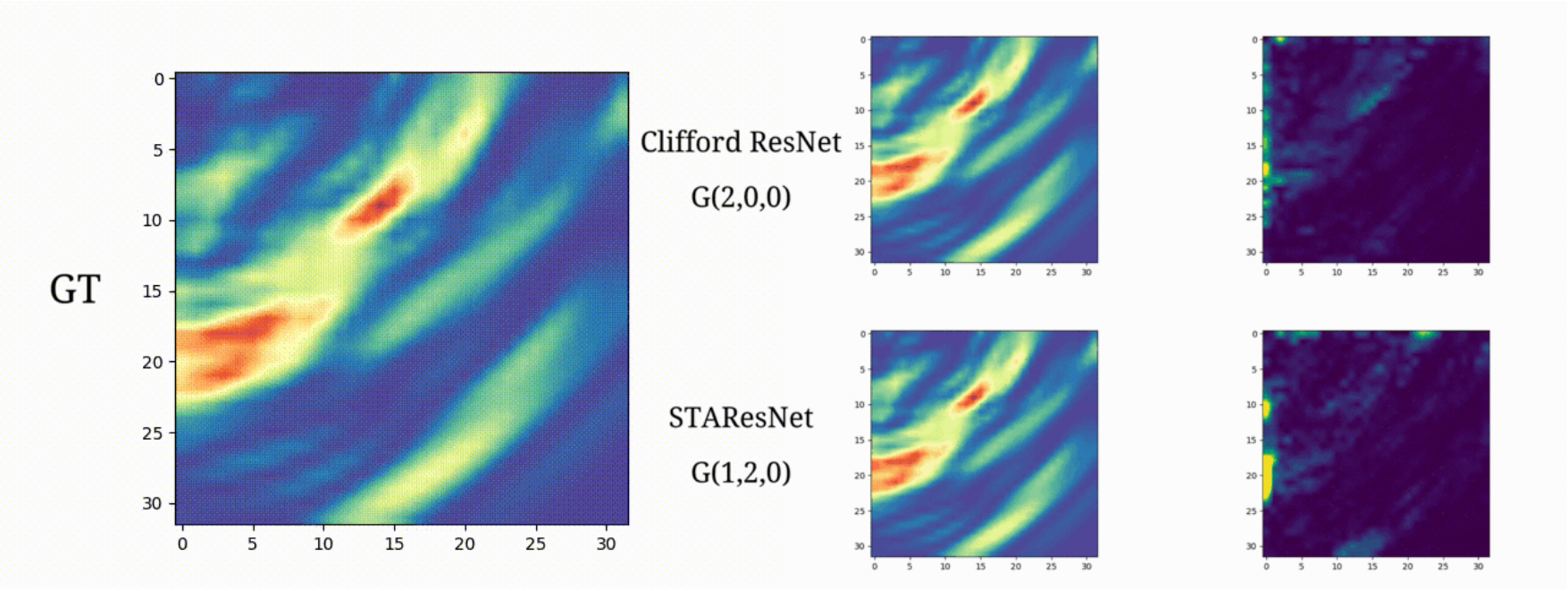}
            \caption[Network2]%
            {{\small $m$ = 1}}    
            \label{fig:meq1new}
        \end{subfigure}
        \hfill
        \begin{subfigure}[b]{0.475\textwidth}  
            \centering 
            \includegraphics[width=\textwidth]{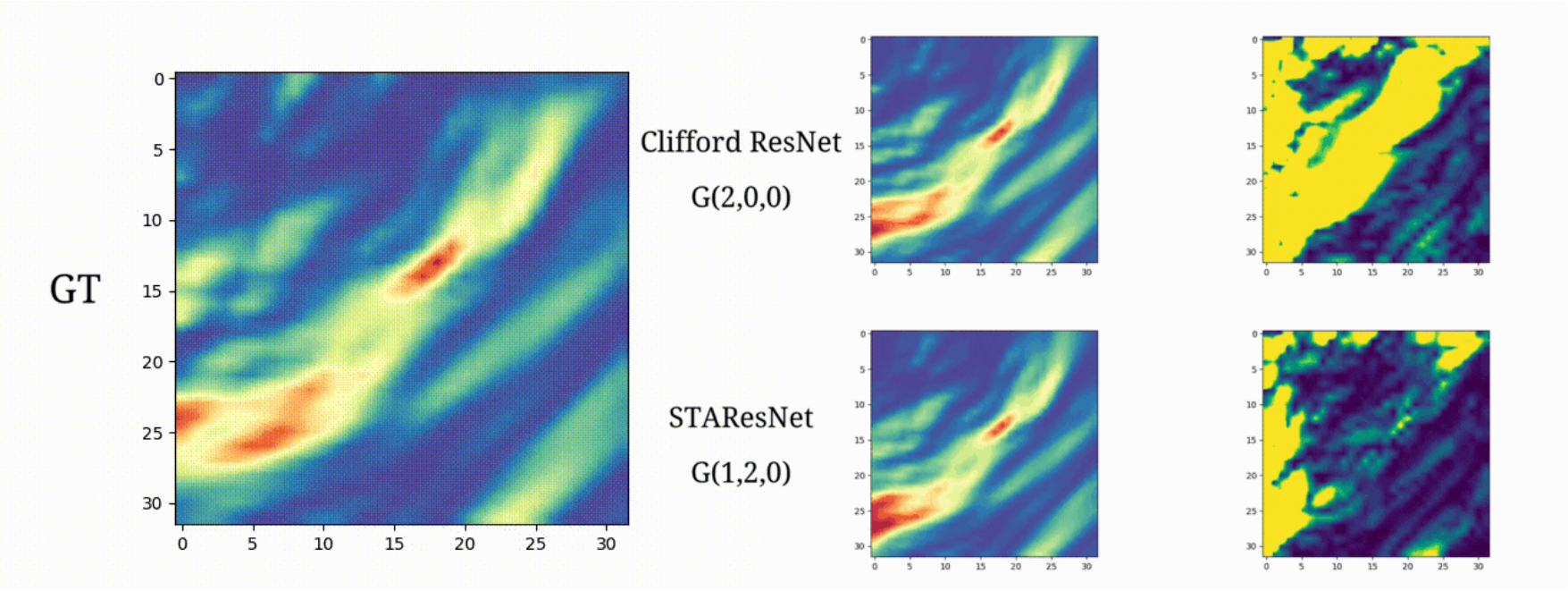}
            \caption[]%
            {{\small $m$ = 5}}    
            \label{fig:meq5new}
        \end{subfigure}
        \vskip\baselineskip
        \begin{subfigure}[b]{0.475\textwidth}   
            \centering 
            \includegraphics[width=\textwidth]{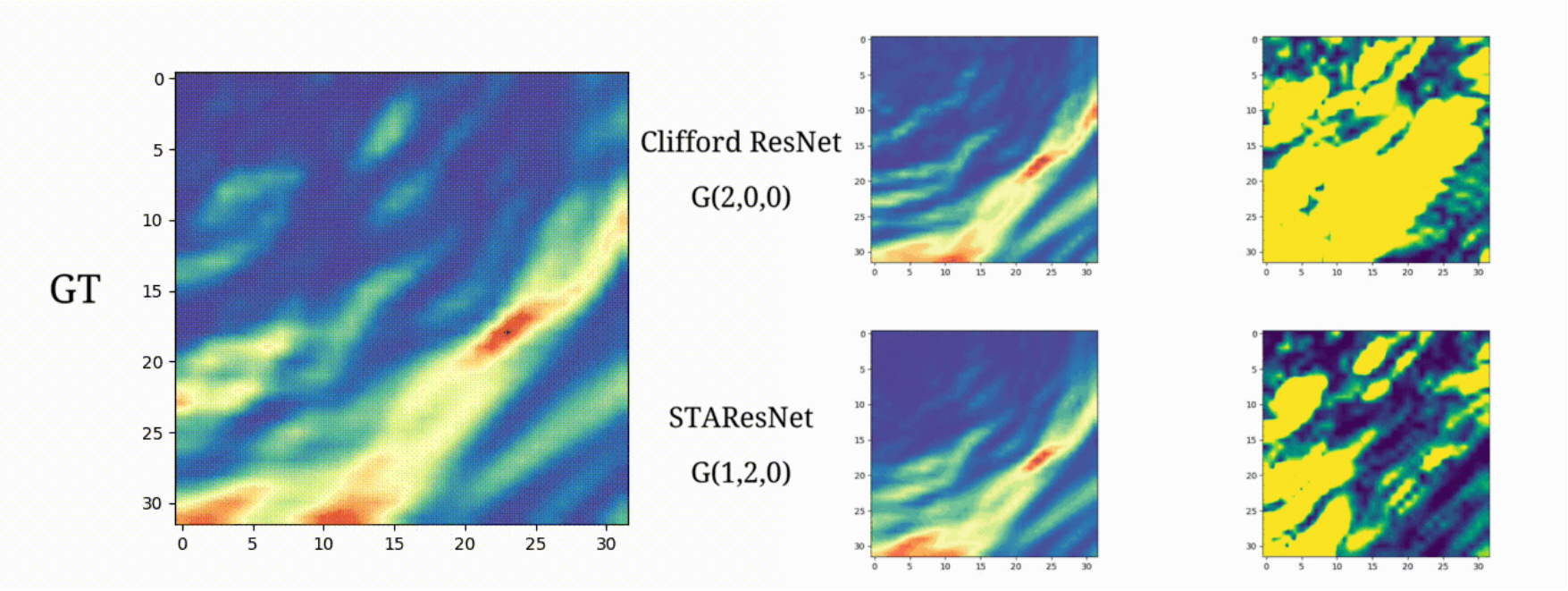}
            \caption[]%
            {{\small $m$ = 8}} 
            \label{fig:meq8new}
        \end{subfigure}
        \hfill
        \begin{subfigure}[b]{0.475\textwidth}   
            \centering 
            \includegraphics[width=\textwidth]{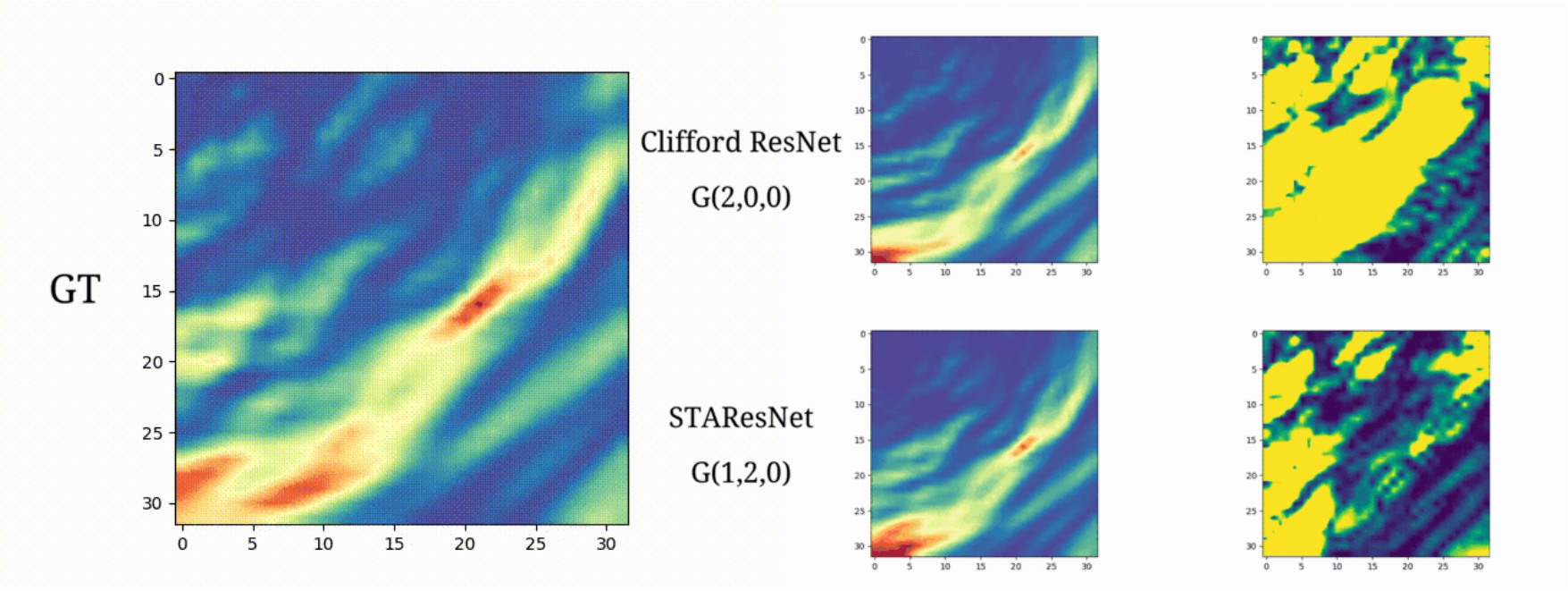}
            \caption[]%
            {{\small $m$ = 10}}    
            \label{fig:meq10new}
        \end{subfigure}
        \caption[ ]
        {\small 2D GT $\mathbf{F}^2$, estimated $\hat{\mathbf{F}}^2$ and difference $|\mathbf{F}^2 - \hat{\mathbf{F}}^2|$ for Clifford ResNet and STAResNet at different rollout steps: (a) $m=1$, (b) $m=5$, (c) $m=8$ and (d) $m=10$. Here $\Delta t = 8$s.} 
        \label{fig:rolloutexamples2}
    \end{figure*}

Examples of the evolution of $\mathbf{F}^2$ as a function of time are given in Figs. \ref{fig:rolloutexamples}-\ref{fig:rolloutexamples2} for two different sequences. We use the $viridis$ colormap is reported the absolute difference, clipped between $[0, 0.02]$, between $\mathbf{F}^2$ and $\hat{\mathbf{F}^2}$. It can be seen how errors in the estimates obtained via STAResNet are visibly more localised and their magnitude more bounded as opposed to those obtained via Clifford ResNet.

\section{Experiments in 3D}

In the 3D case we consider a volume with spatial resolution of $28 \times 28 \times 28$ and step size $\Delta x = \Delta y = \Delta z =  5\cdot 10^{-7}$m, with the EM field sampled with varying sampling period $\Delta t = \{5, 8, 10, 15\}$s. The light is propagated from 6 point sources randomly placed in each of the the $xy$, $yz$ and $xz$ planes, for a total of 18 planar sources.  The wavelength of the emitted light is $\lambda = 10^{-5}m$. Each light source emits light with a random phase and random amplitude. Training, validation and test sets are structured similarly to their 2D counterparts.

We measure the MSE error between estimated and GT fields as described in Eq. \ref{loss3D}, and visualise the magnitude of the 4D spacetime bivector:  \begin{equation}
    \mathbf{F}^2 = (E_1 \gamma_{10} +E_2 \gamma_{20} +E_3 \gamma_{30} + B_1 \gamma_{13} + B_2 \gamma_{13} + B_3 \gamma_{12})^2 
    \label{faradaybiv3d}
\end{equation}

\begin{figure*}[!htpb]
        \centering
        \begin{subfigure}[b]{0.45\textwidth}
            \centering
            \includegraphics[width=\textwidth]{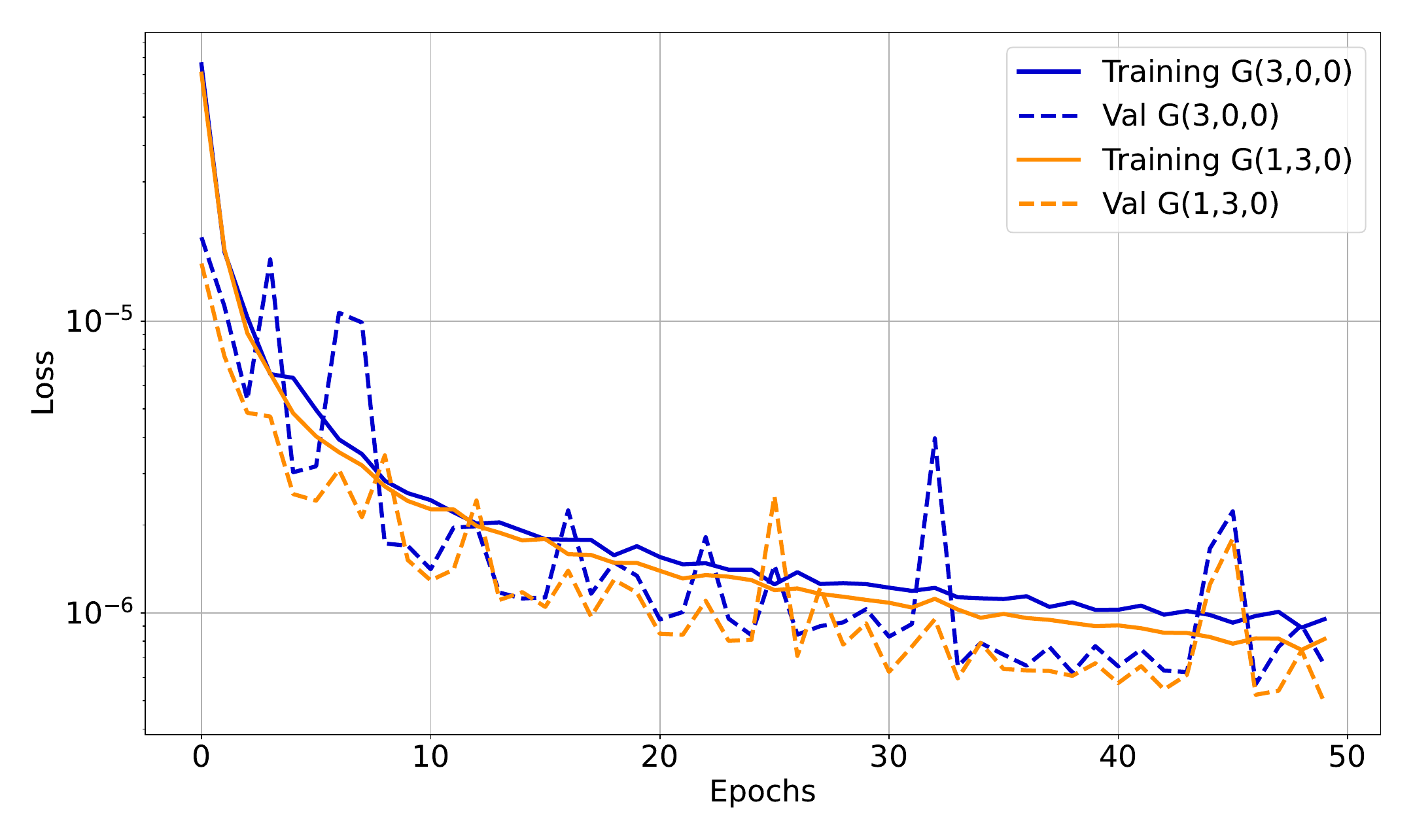}
            \caption[Network2]%
            {{\small $\Delta t = 5$s}}    
            \label{fig:loss5s}
        \end{subfigure}
        \hfill
        \begin{subfigure}[b]{0.45\textwidth}  
            \centering 
            \includegraphics[width=\textwidth]{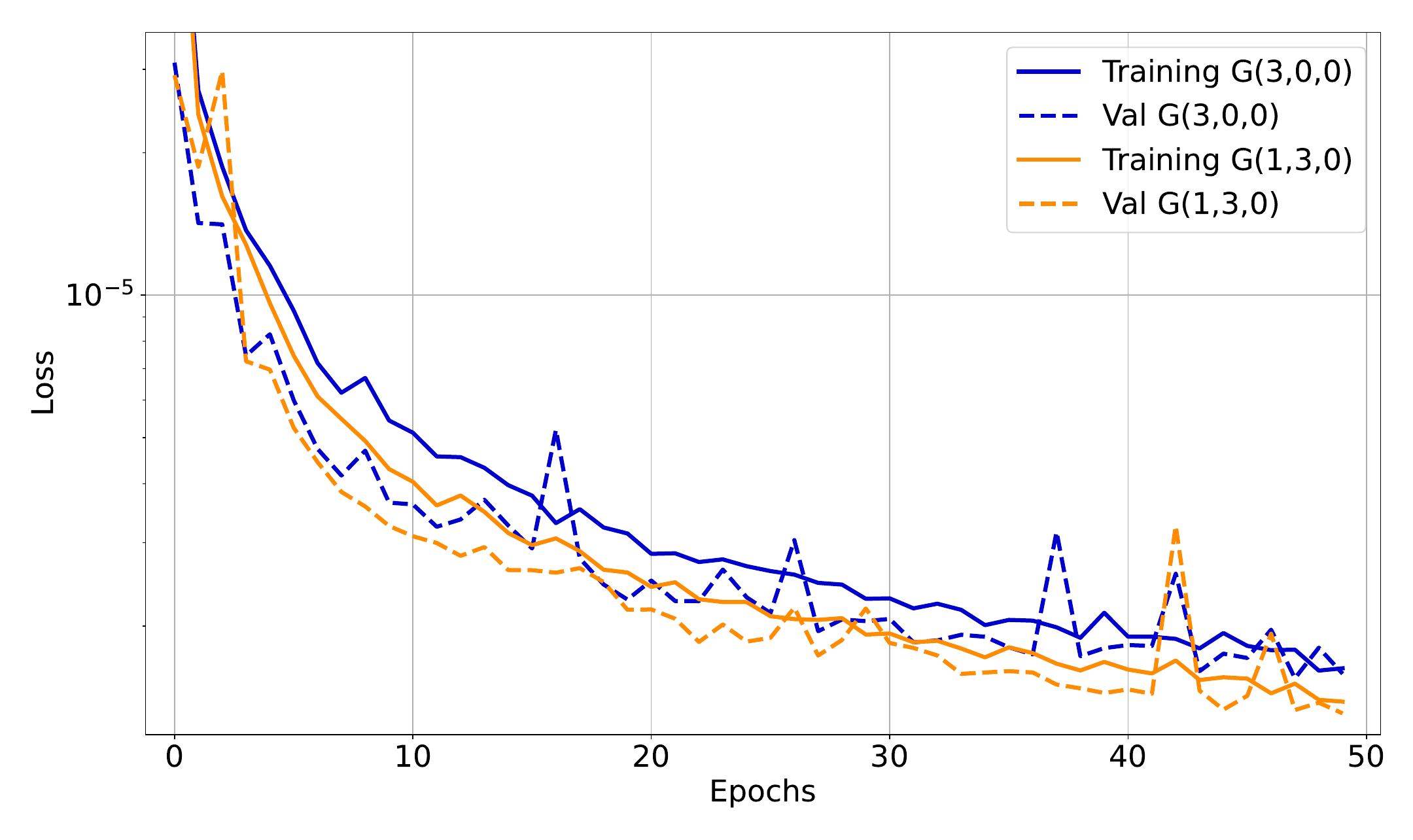}
            \caption[]%
            {{\small $\Delta t = 8$s}}    
            \label{fig:loss8s}
        \end{subfigure}
        \vskip\baselineskip
        \begin{subfigure}[b]{0.45\textwidth}   
            \centering 
            \includegraphics[width=\textwidth]{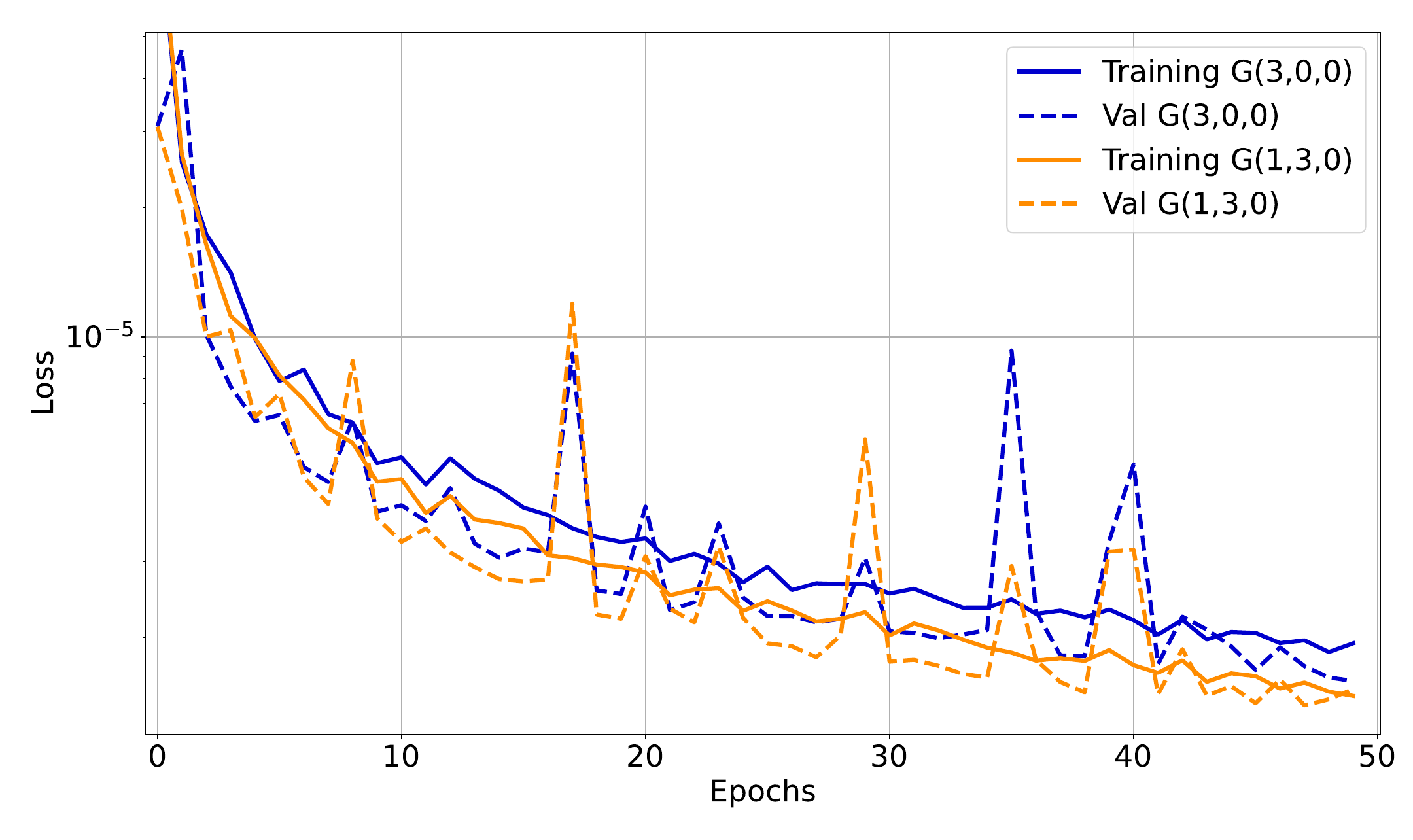}
            \caption[]%
            {{\small $\Delta t = 10$s}} 
            \label{fig:loss10s}
        \end{subfigure}
        \hfill
        \begin{subfigure}[b]{0.45\textwidth}   
            \centering 
            \includegraphics[width=\textwidth]{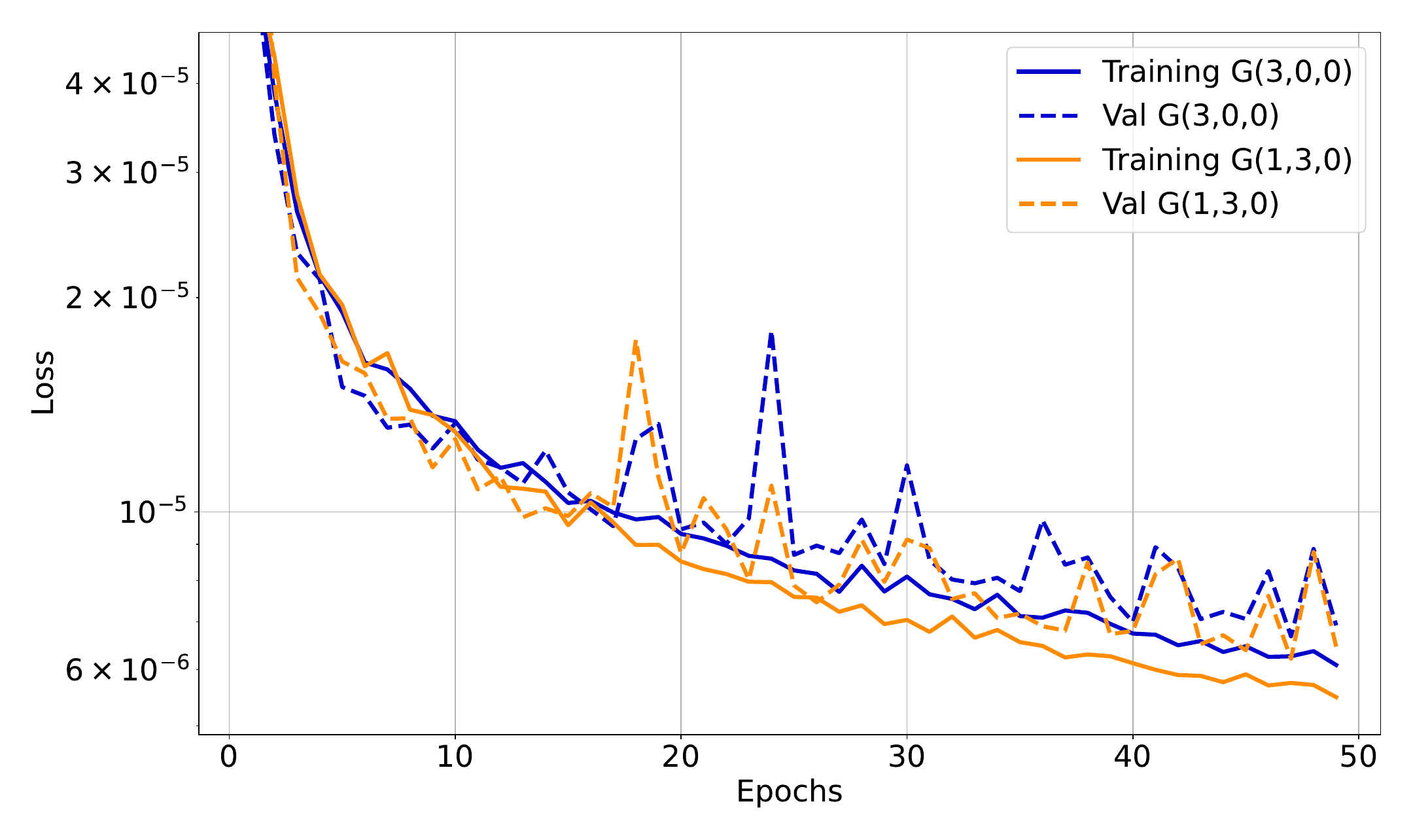}
            \caption[]%
            {{\small $\Delta t = 15$s}}    
            \label{fig:loss15s}
        \end{subfigure}
        \caption[ ]
        {\small Training and validation losses versus number of epochs for 2D Maxwell's PDEs for instances sampled at (a) $5$s, (b) $8$s, (c) $10$s, (d) $15$s.} 
        \label{fig:losses3D}
    \end{figure*}

Training and validation losses for Clifford ResNet and STAResNet at different sampling periods are shown in Fig. \ref{fig:losses3D}. Our STA approach achieves lower validation loss in all four cases, similarly to Fig. \ref{fig:losses}. We believe that the smaller gap between the loss profiles of STAResNet and CliffordResNet, as opposed to that in Fig. \ref{fig:losses}, is to be attributed to the small $\Delta t$ at which the dataset is generated and smaller size of the domain rather than the fact that we are working in 3D over 2D. 

\begin{figure}[!htbp]
    \centering
    \includegraphics[width=0.6\textwidth]{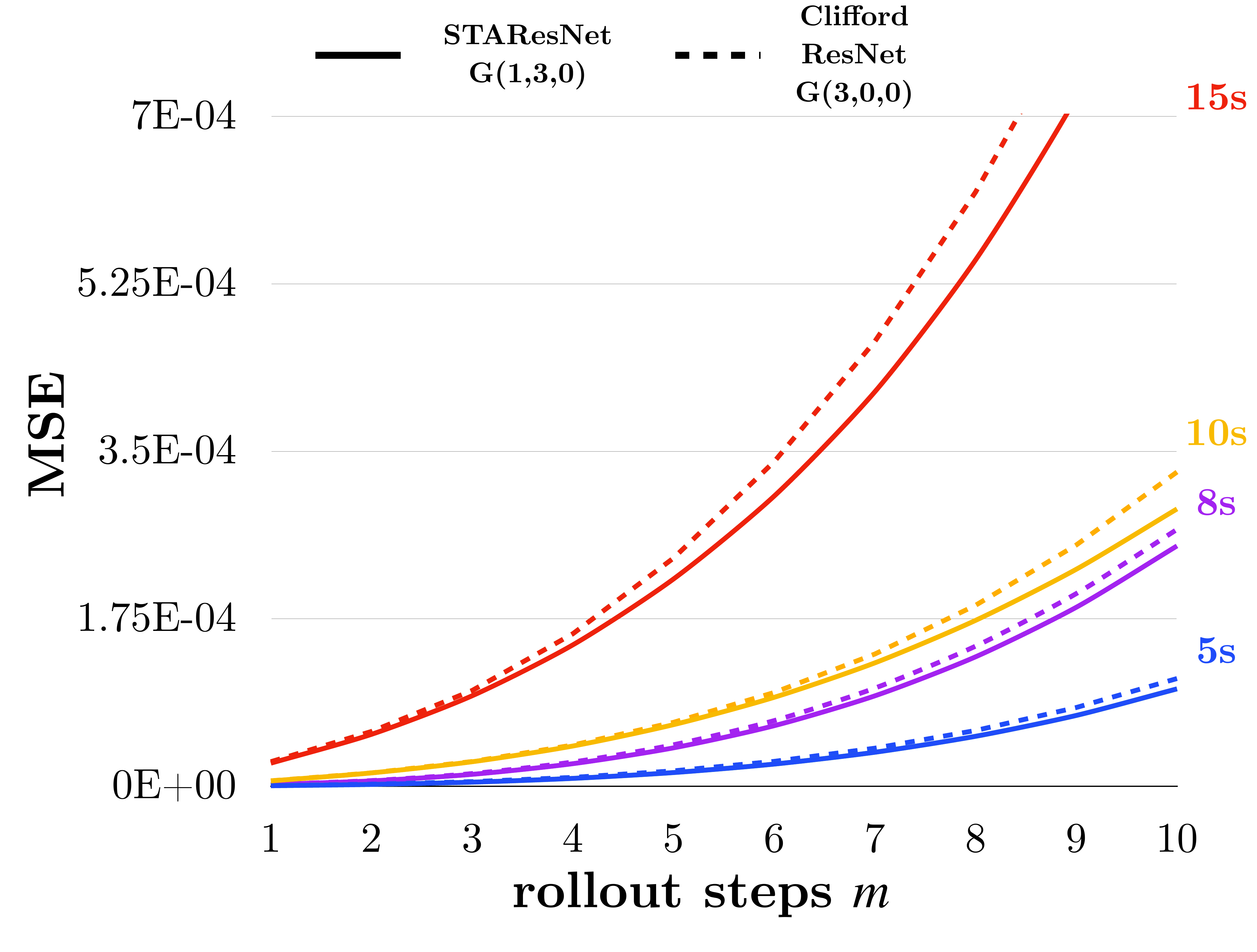}
    \caption{Mean squared error between estimated and ground truth EM fields over test set versus rollout steps $m$ for the 3D case.}
    \label{fig:rollout3D}
\end{figure}

A plot of the MSE as a function of the number of rollout steps $m$ for successive time instants is given in Fig. \ref{fig:rollout3D}. For each of the four sampling periods and regardless of the number of rollout steps, STAResNet consistently outperforms Clifford ResNet. Notice how this holds true also for $\Delta t = 15$s, while in the 2D case, in Fig. \ref{fig:rollout}, errors explode for $\Delta t \geq 15$s.

\begin{figure*}[!htbp]
        \centering
               \begin{subfigure}[b]{0.475\textwidth}
            \centering
            \includegraphics[width=\textwidth]{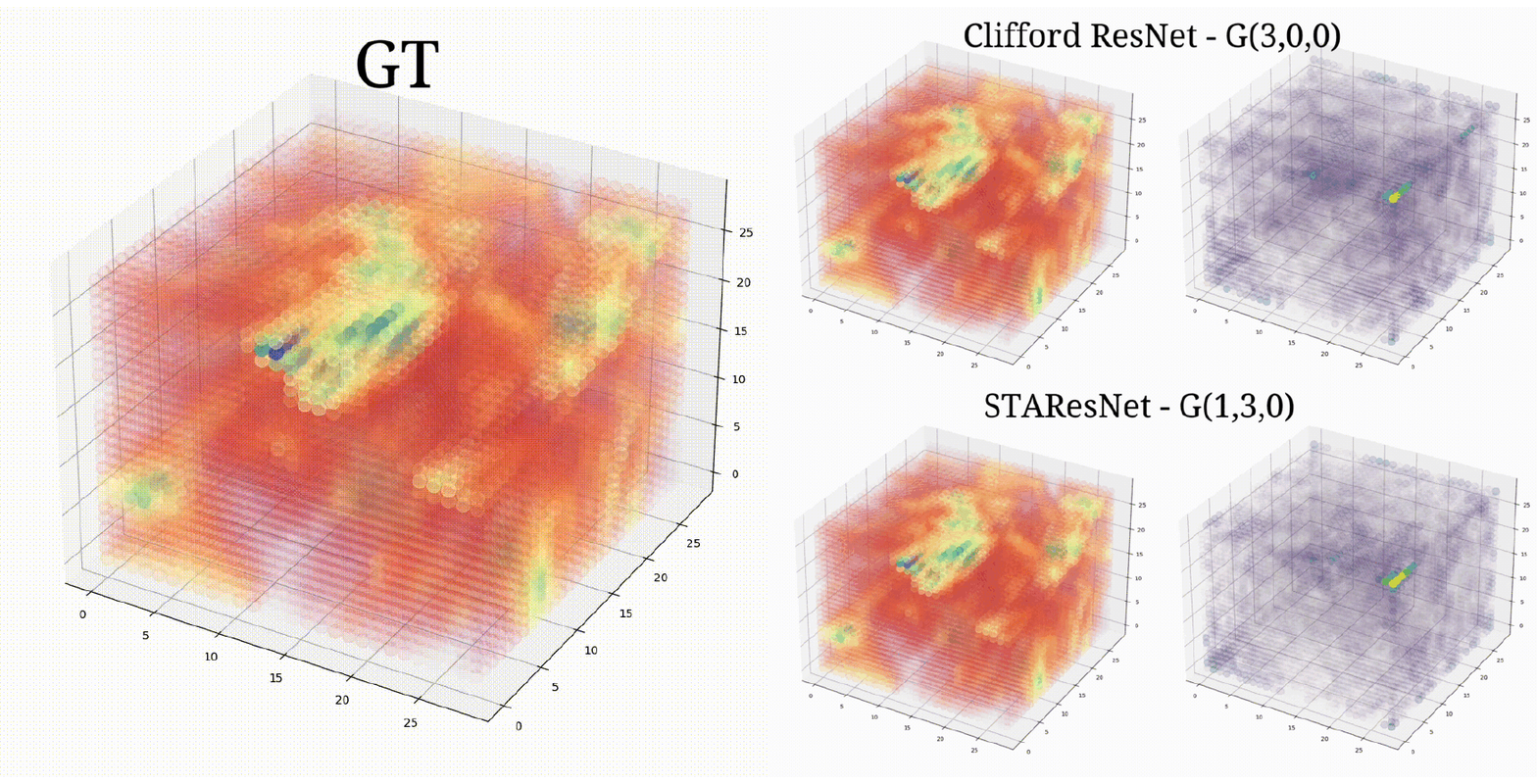}
            \caption[Network2]%
            {{\small $m$ = 2}}    
            \label{fig:ex1}
        \end{subfigure}
        \hfill
        \begin{subfigure}[b]{0.475\textwidth}   
            \centering 
            \includegraphics[width=\textwidth]{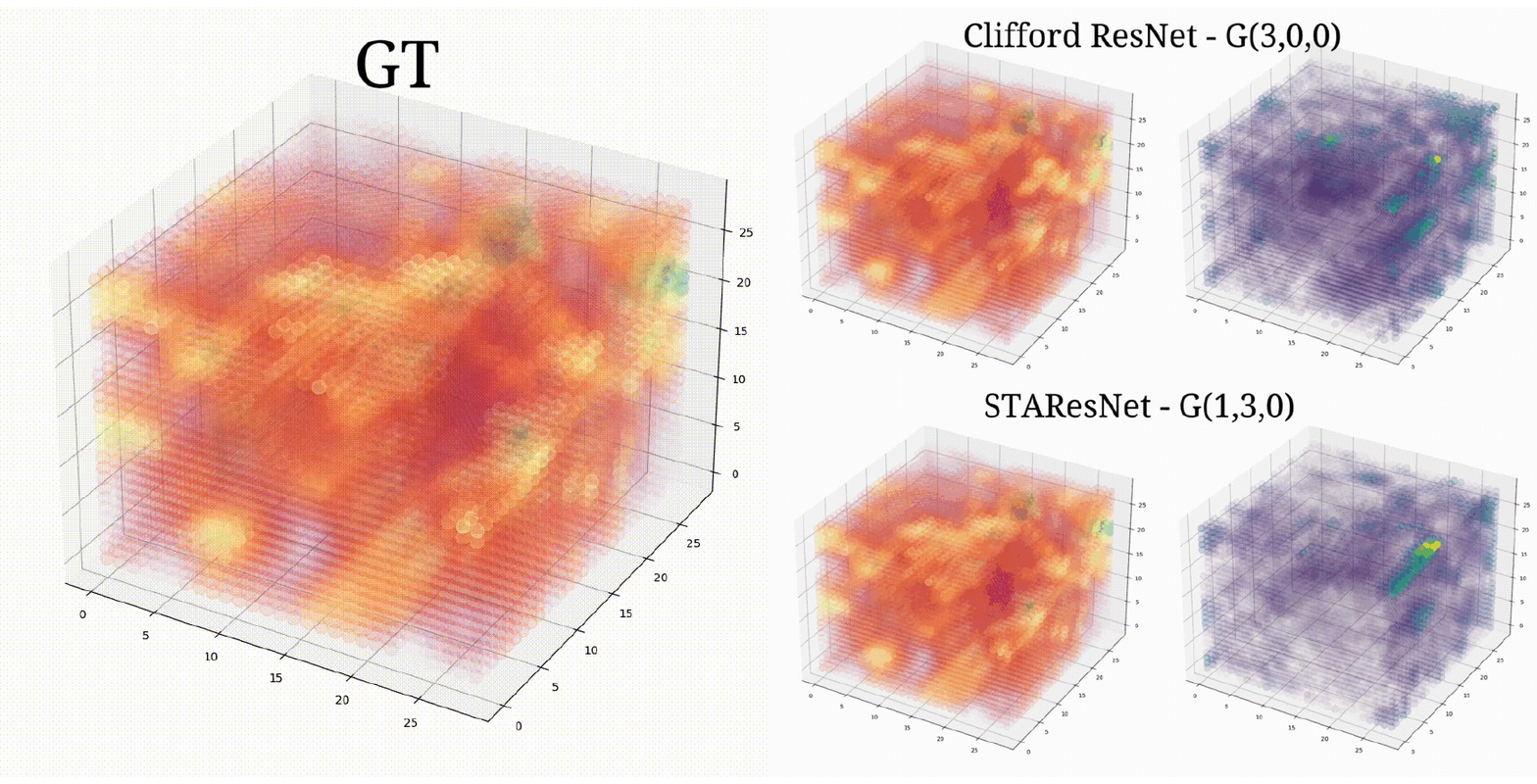}
            \caption[]%
            {{\small $m$ = 5}}    
            \label{fig:ex00}
        \end{subfigure}
        \begin{subfigure}[b]{0.475\textwidth}
            \centering
            \includegraphics[width=\textwidth]{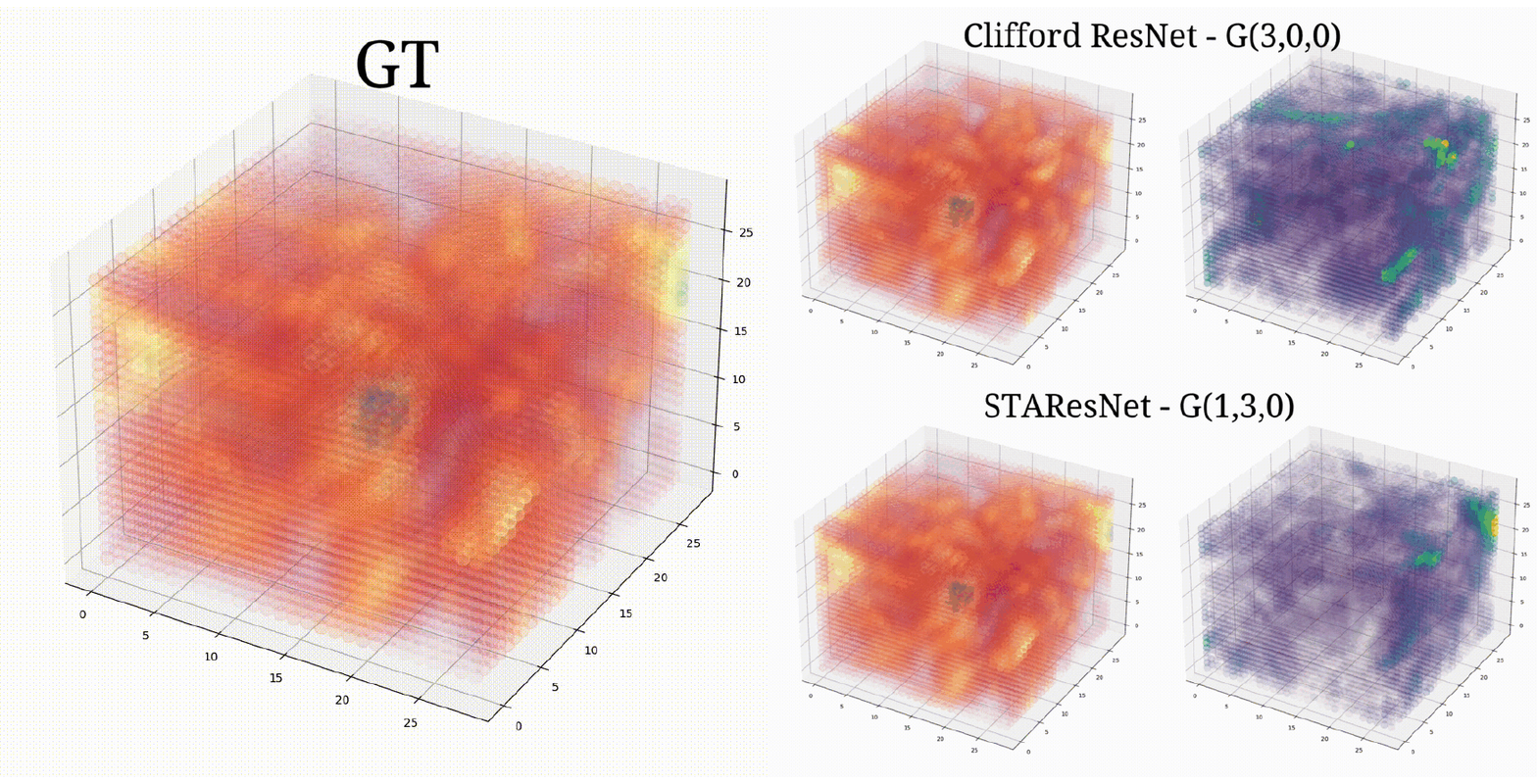}
            \caption[Network2]%
            {{\small $m$ = 7}}    
            \label{fig:ex2}
        \end{subfigure}
        \hfill
        \begin{subfigure}[b]{0.475\textwidth}   
            \centering 
            \includegraphics[width=\textwidth]{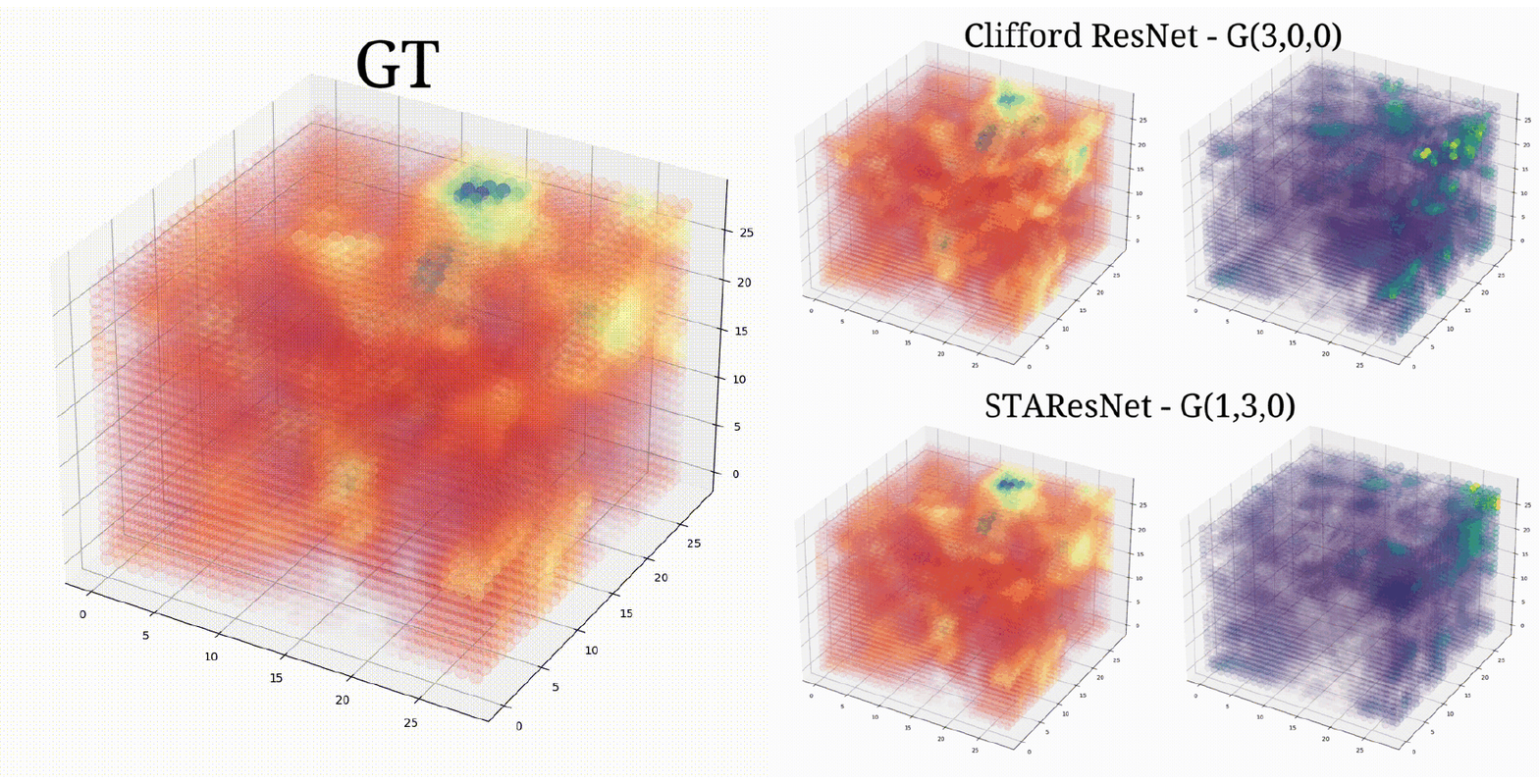}
            \caption[]%
            {{\small $m$ = 10}}    
            \label{fig:ex3}
        \end{subfigure}
        \caption[ ]
        {\small 2D GT $\mathbf{F}^2$, estimated $\hat{\mathbf{F}}^2$ and difference $|\mathbf{F}^2 - \hat{\mathbf{F}}^2|$ for Clifford ResNet and STAResNet at different rollout steps: (a) $m=2$, (b) $m=5$, (c) $m=7$ and (d) $m=10$. Here $\Delta t = 5$s. Transparency is proportional to intensity.} 
        \label{fig:rolloutexamples3d}
    \end{figure*}

Examples of the evolution of the magnitude of the 3D Faraday bivector for successive rollout steps are shown in Fig. \ref{fig:rolloutexamples3d}. The absolute difference between GT and estimated 3D Faraday bivectors' magnitudes clipped between $[0, 0.02]$ is shown in $viridis$ color map, meaning they share the same range. The higher the intensity and the opacity of the plot, the higher the error between ground truth. It can be noticed how, even after only two time steps, STAResNet yields output closer to GT as opposed to the 3D Clifford ResNet. 

\begin{figure*}[!htbp]
        \centering
               \begin{subfigure}[b]{0.8\textwidth}
            \centering
            \includegraphics[width=\textwidth]{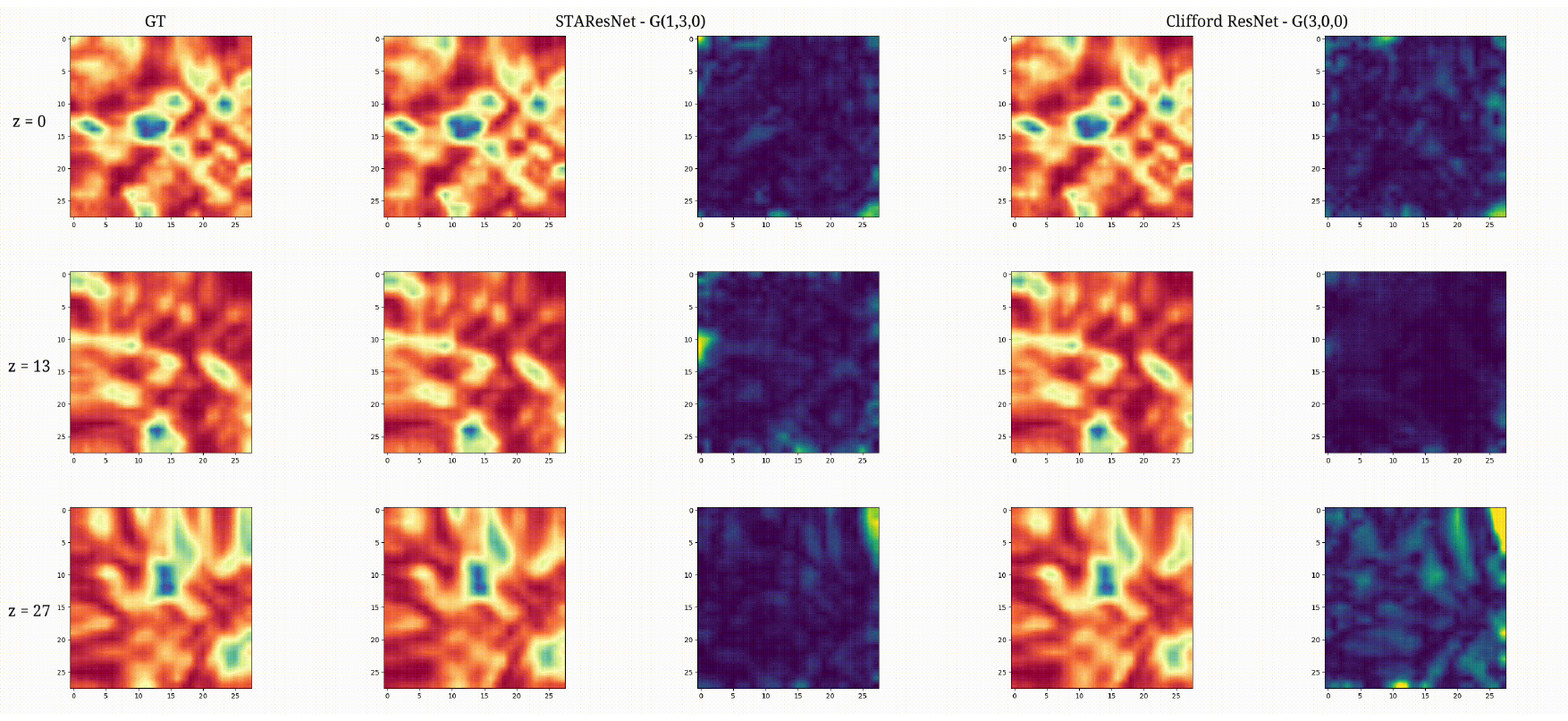}
            \caption[Network2]%
            {{\small $m$ = 3}}    
            \label{fig:ex23}
        \end{subfigure}
        \vspace{0pt} 
        \begin{subfigure}[b]{\textwidth}   
            \centering 
            \includegraphics[width=0.8\textwidth]{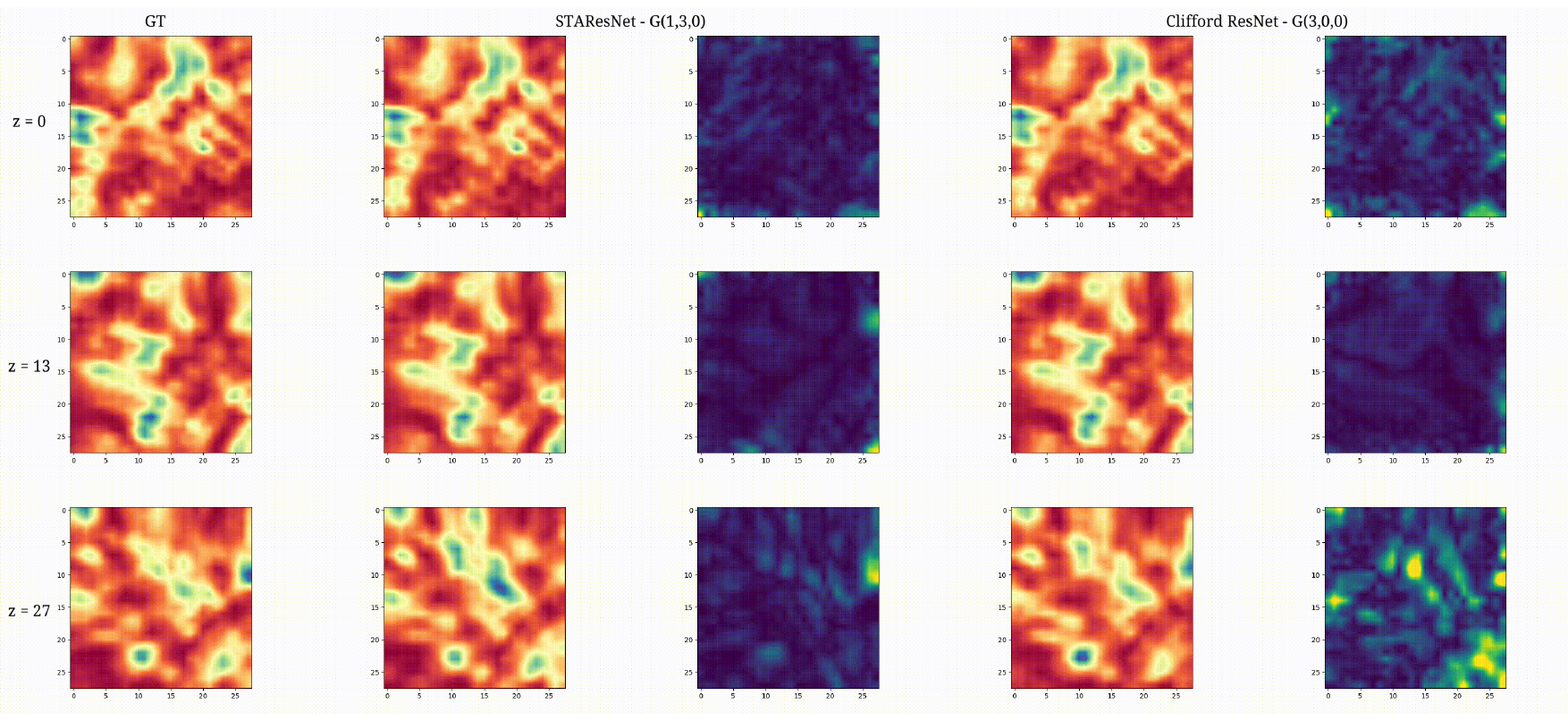}
            \caption[]%
            {{\small $m$ = 5}}    
            \label{fig:ex0}
        \end{subfigure}
        \caption[ ]
        {\small Slices of the 3D GT $\mathbf{F}^2$, estimated $\hat{\mathbf{F}}^2$ and difference $|\mathbf{F}^2 - \hat{\mathbf{F}}^2|$ at different heights $z$ of the volume for Clifford ResNet and STAResNet at different rollout steps: (a) $m = 3$, (b) $m =5$. $\Delta t = 5$s.} 
        \label{fig:rolloutexamplesslices}
    \end{figure*}

\begin{figure*}[!htbp]
\centering
        \begin{subfigure}[b]{\textwidth}
            \centering
            \includegraphics[width=0.8\textwidth]{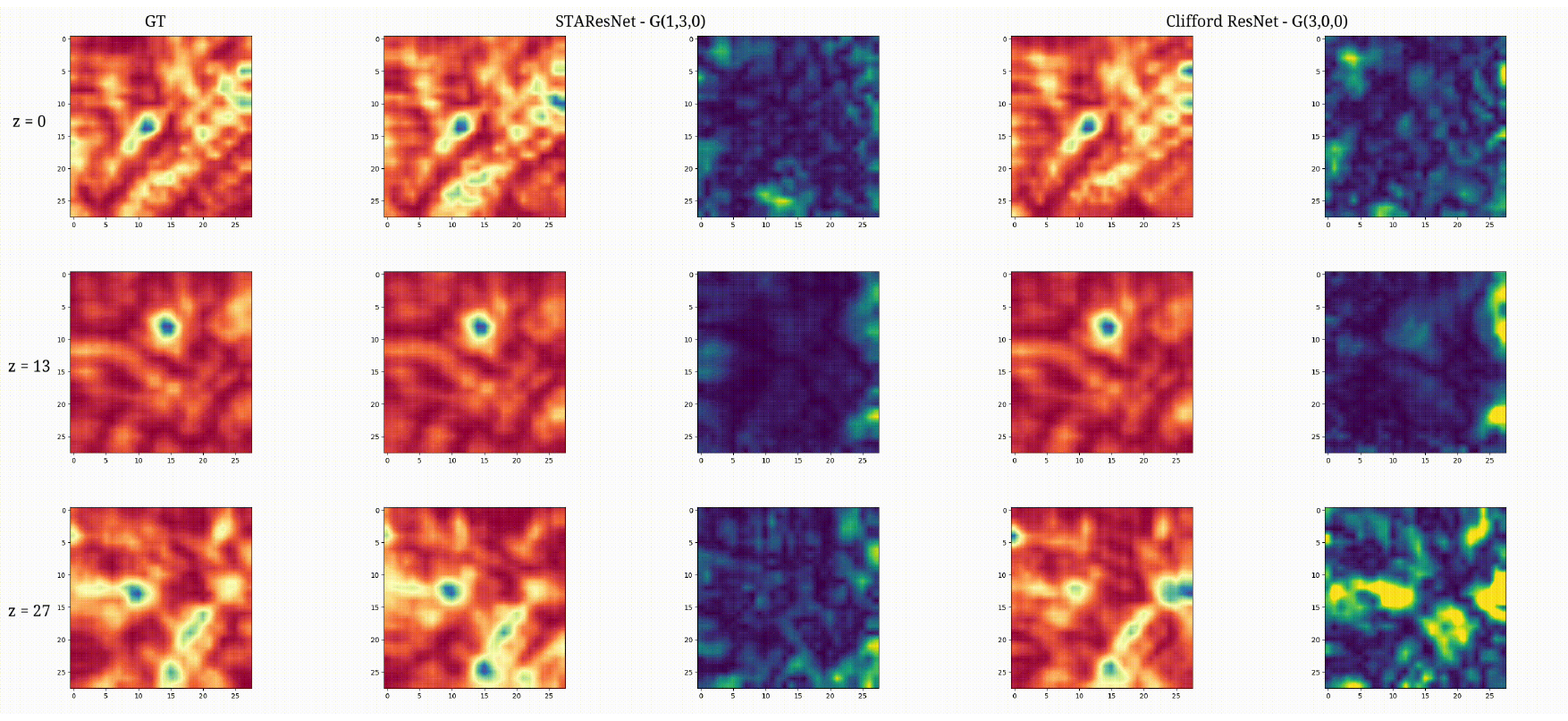}
            \caption[Network2]%
            {{\small $m$ = 8}}    
            \label{fig:ex22}
        \end{subfigure}
        \hspace{1em}
        \begin{subfigure}[b]{\textwidth}   
            \centering 
            \includegraphics[width=0.8\textwidth]{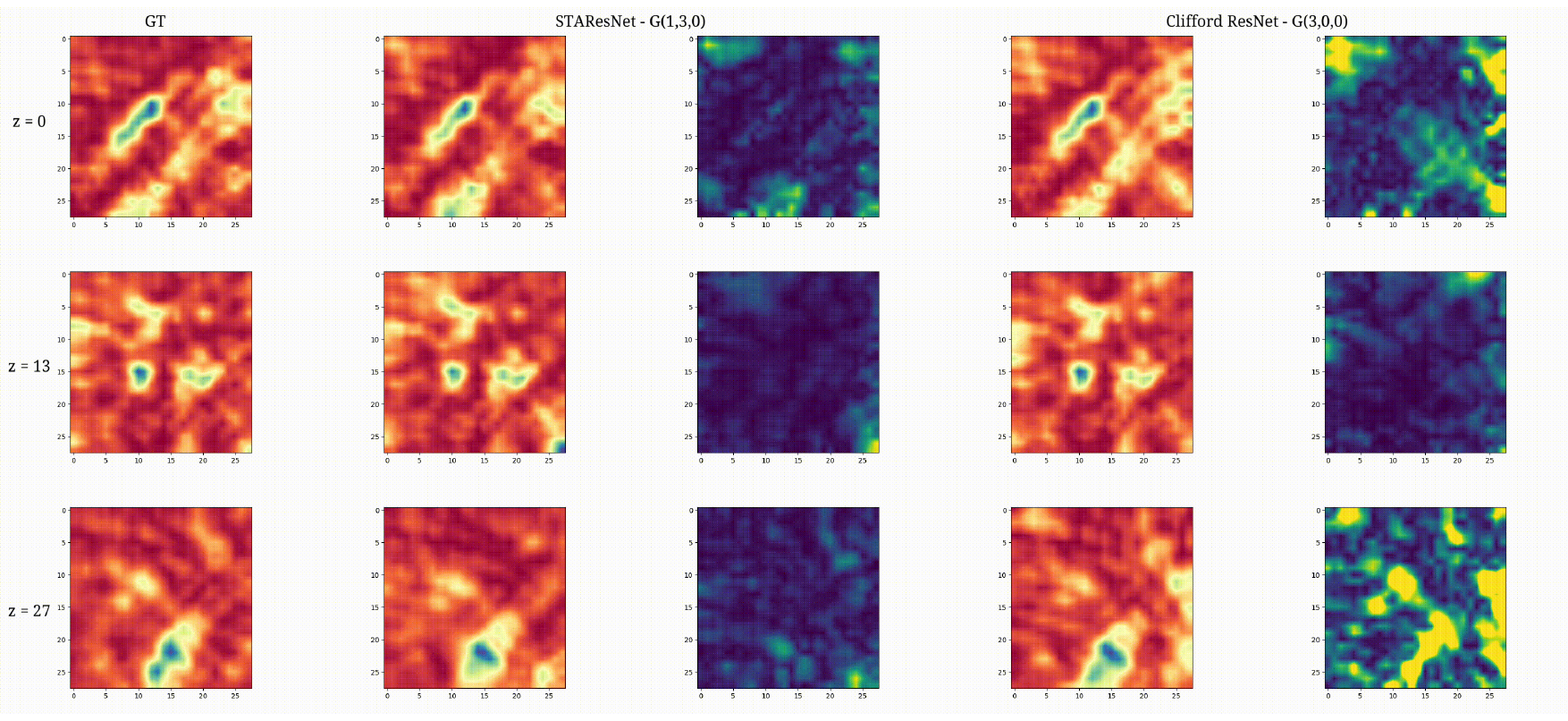}
            \caption[]%
            {{\small $m$ = 10}}    
            \label{fig:ex33}
        \end{subfigure}
        \caption[ ]
        {\small Slices of the 3D GT $\mathbf{F}^2$, estimated $\hat{\mathbf{F}}^2$ and difference $|\mathbf{F}^2 - \hat{\mathbf{F}}^2|$ at different heights $z$ of the volume for Clifford ResNet and STAResNet at different rollout steps: (a) $m = 8$, (b) $m =10$. $\Delta t = 5$s.} 
        \label{fig:rolloutexamplesslices2}
    \end{figure*}

Similarly, slices along the $z$-dimension of the 3D Faraday bivector for successive rollout steps are shown in Fig. \ref{fig:rolloutexamplesslices}-\ref{fig:rolloutexamplesslices2}. Slices are taken at the bottom ($z = 0$), middle ($z = 13$) and top ($z = 27$) of the volume. The clipped, absolute differences are shown in $viridis$ color map. Clifford ResNet yields significantly larger errors for each of the four rollout steps and across each of the three vertical slices considered.

\section{Conclusions \& Future Work}
We have introduced STAResNet, a ResNet-inspired architecture that works with multivectors in STA, to shed light on the importance of choosing the right mathematical space for GA networks. We compared the performance of STAResNet, which is the first example of a network working in spacetime and with STA multivectors, with Clifford ResNet, working in vanilla GA, on the solution of Maxwell's equations.

It had already been demonstrated in the literature that a description of Maxwell's PDEs in STA offers an easier way to solve them numerically \cite{lasenby1998gravity}. We verified that this holds true also when the solution of Maxwell's PDEs is learnt, as long as the learning happens in STA. 

STAResNet outperforms Clifford ResNet in 2D and 3D, at different sampling periods, in the presence of obstacles either in a previously seen or unseen configuration, and over multiple time steps into the future. Most notably, STAResNet is able to generalise better over previously unseen data and achieve a lower error at a fraction of the number of trainable parameters as opposed to Clifford ResNet.

This substantial improvement is not surprising as it is grounded in the physics of the problem: working with Faraday bivectors in STA is a more natural and appropriate choice over the modelling of the EM fields as mixed-grade multivectors with vector and bivector components describing electric and magnetic fields, respectively. 

We therefore conclude that the choice of the right algebra in Clifford networks is a key factor in obtaining more descriptive, compact and accurate learning pipelines.

Future works to further corroborate our thesis might include the study of different PDEs that can be expressed in more than one algebra and solving them with Clifford networks in those algebras.

\bibliographystyle{unsrt}  
\bibliography{references}

\end{document}